\documentclass[letterpaper, 10 pt, journal, twoside]{ieeetran}  

\usepackage{amssymb,graphicx,amsmath,color,algorithm,algorithmic,url}
\usepackage[english]{babel}
\usepackage{blindtext}
\usepackage{afterpage}
\usepackage[noadjust]{cite} 
\usepackage{hhline,caption}
\usepackage{hyperref}
\IEEEoverridecommandlockouts                              

\title{\LARGE \bf
\color{black}{DigiTac: A DIGIT-TacTip Hybrid Tactile Sensor for Comparing\\ Low-Cost High-Resolution Robot Touch}
\vspace{0em}
}

\author{Nathan F. Lepora, Yijiong Lin, Ben Money-Coomes, John Lloyd\\
\url{www.lepora.com/digitac}, \url{www.github.com/nlepora/digitac-design}\vspace{-1.5em}%
\thanks{NL and JL were supported by a Leadership Award from the Leverhulme Trust on ‘A biomimetic forebrain for robot touch’ (RL-2016-39).}
\thanks{All authors are with the Department of Engineering Mathematics and Bristol Robotics Laboratory, University of Bristol, Bristol, U.K.}
\thanks{Corresponding author: n.lepora@bristol.ac.uk}
\thanks{Project website: \url{www.lepora.com/digitac}}
\thanks{Project repository: \url{www.github.com/nlepora/digitac-design}}
\thanks{Shared data: \url{www.doi.org/10.5523/bris.110f0tkyy28pa2joru2pxxbrxd}}
}
\begin{document}
\maketitle

\begin{abstract}
Deep learning combined with high-resolution tactile sensing could lead to highly capable dexterous robots. However, progress is slow because of the specialist equipment and expertise. The DIGIT tactile sensor offers low-cost entry to high-resolution touch using GelSight-type sensors. Here we customize the DIGIT to have a 3D-printed sensing surface based on the TacTip family of soft biomimetic optical tactile sensors. The DIGIT-TacTip (DigiTac) enables direct comparison between these distinct tactile sensor types. For this comparison, we introduce a tactile robot system comprising a desktop arm, mounts and 3D-printed test objects. We use tactile servo control with a PoseNet deep learning model to compare the DIGIT, DigiTac and TacTip for edge- and surface-following over 3D-shapes. All three sensors performed similarly at pose prediction, but their constructions led to differing performances at servo control, offering guidance for researchers selecting or innovating tactile sensors. All hardware and software for reproducing this study will be openly released.
\end{abstract}

\begin{IEEEkeywords} Force and Tactile Sensing, Open-source Robotics \end{IEEEkeywords}

\section{INTRODUCTION}

Advances in deep learning combined with innovation in high-resolution tactile sensing give a plausible route to robots with the dexterous capabilities of the human hand. Consider, for example, the revolution in computer vision where 1000s of researchers have ready access to deep learning infrastructure and vast amounts of openly-shared visual data. In contrast, research on robot touch for dexterity requires access to specialist equipment such as industrial arms integrated with tactile sensing hardware. This need for specialised equipment and expertise is a barrier to entering the field and reproducing research findings. Consequently, progress is slow and narrow in scope from just a few research laboratories.

Researchers in Meta AI have developed the DIGIT as a low-cost high-resolution tactile sensor~\cite{lambeta_digit_2020-2}. They have open-sourced the sensor design and in partnership with the company GelSight commercially manufacture the sensor~\cite{digit_notitle_nodate}. To lower the barrier of entry to using machine learning (ML) with tactile sensors like the DIGIT, an accompanying library of ML models and functionality (PyTouch) was also released~\cite{lambeta_pytouch_2021} alongside a simulation environment (TACTO)~\cite{wang_tacto_2022}. The overall concept is to provide a complete ecosystem for teaching robots to perceive, understand and interact through touch~\cite{noauthor_teaching_nodate}.

\textcolor{black}{For low-cost high-resolution robot touch to progress, there needs to be a comparison of tactile sensing technologies and a greater diversity of tactile sensing hardware, ML models and test scenarios. The TacTip is a distinct high-resolution optical tactile sensor with a 3D-printed biomimetic structure of papillae that amplify the movement of markers on their tips~\cite{ward-cherrier_tactip_2018-1,lepora_soft_2021-1}, which also has body of research on using ML models for tactile perception and control. In this paper, we customize the DIGIT to have a TacTip skin to make the first direct comparison between the GelSight and TacTip families of tactile sensors using common software and hardware. For this comparison, we make use of recent progress in pose-based servo control using ML models on tactile images~\cite{lepora_pixels_2019-1,lepora_optimal_2020-1,lepora_pose-based_2021}.}
 
\begin{figure}[t!]
	\centering
	\begin{tabular}[b]{@{}c@{}}
	    {\bf Test System: Low-Cost Tactile Sensor on Desktop Arm}\\
        \includegraphics[width=\columnwidth,trim={0 0 0 0},clip]{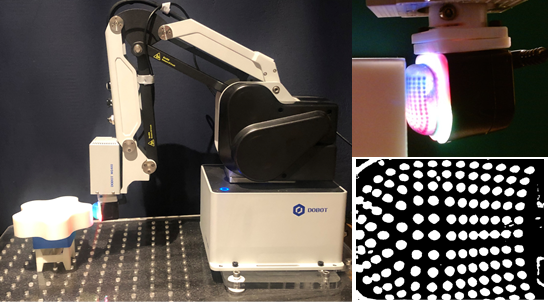} 
    \end{tabular}
	\caption{\textcolor{black}{Left: the DIGIT-TacTip (DigiTac) mounted on an affordable desktop robot arm (Dobot MG400, $\sim$£2k). Right: close-up of sensor and a captured high-resolution tactile image.}}
	\label{fig:1}
	\vspace{-1em}
\end{figure}


Our main contributions are: \\
\textcolor{black}{\noindent 1) We introduce the DIGIT-TacTip (DigiTac) that combines the DIGIT base (lighting, camera board, housing) with a TacTip sensing module (3D-printed skin/mount, window, soft gel). This raises challenges in adapting a design intended for GelSight-type sensors to the TacTip sensing mechanism. In particular, the DIGIT is designed for lateral illumination of a flat sensing surface, whereas the TacTip works best with a curved skin under direct lighting.} \\
\noindent 2) We apply recent developments in pose-based tactile servo control~\cite{lepora_optimal_2020-1,lepora_pose-based_2021} to compare the sensors on a common affordable desktop robot system (Fig.~1), by learning to slide over the edges and surfaces of unknown 3D objects while maintaining safe, delicate contact. This raises several challenges, including how to apply methods developed with industrial robots to lower-capability desktop robots, how to use and compare distinct high-resolution optical tactile sensors with the same hardware/software infrastructure, and how to conduct the experiments in a manner that facilitates reproducible research.   

All hardware and software components including 3D-printable designs for the DigiTac will be openly released. 



\begin{figure*}[t!]
	\centering
	\begin{tabular}[b]{@{}c@{}c@{}c@{}c@{}c@{}}
		{\bf (a) TacTip (round)} & {\bf (b) TacTip (flat)} & {\bf (c) DigiTac} & {\bf (d) DIGIT} & {\bf (e) TacTip (fingertip)} \\
		\includegraphics[width=0.41\columnwidth,trim={450 250 550 100},clip]{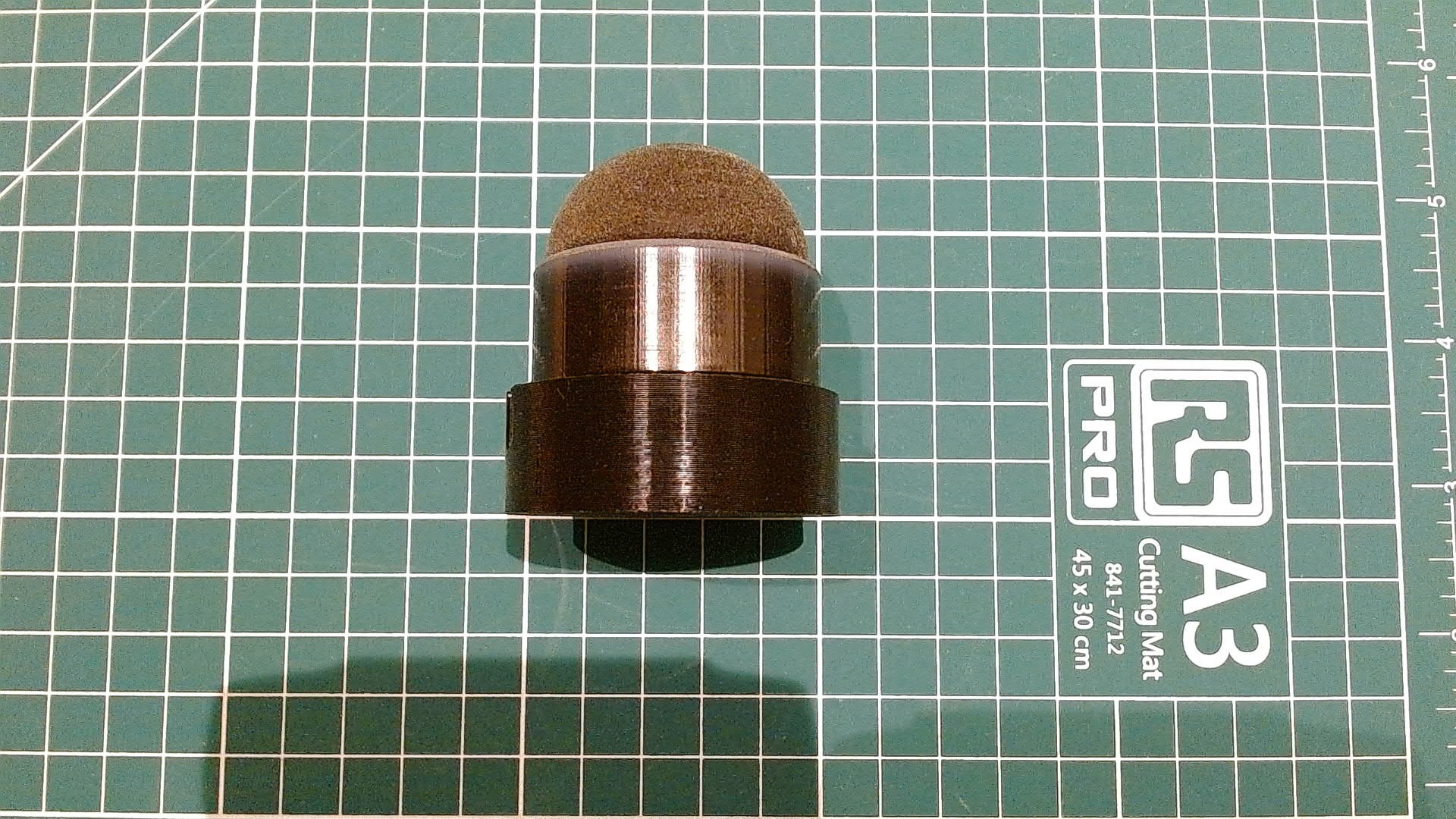} &
		\includegraphics[width=0.41\columnwidth,trim={450 250 550 100},clip]{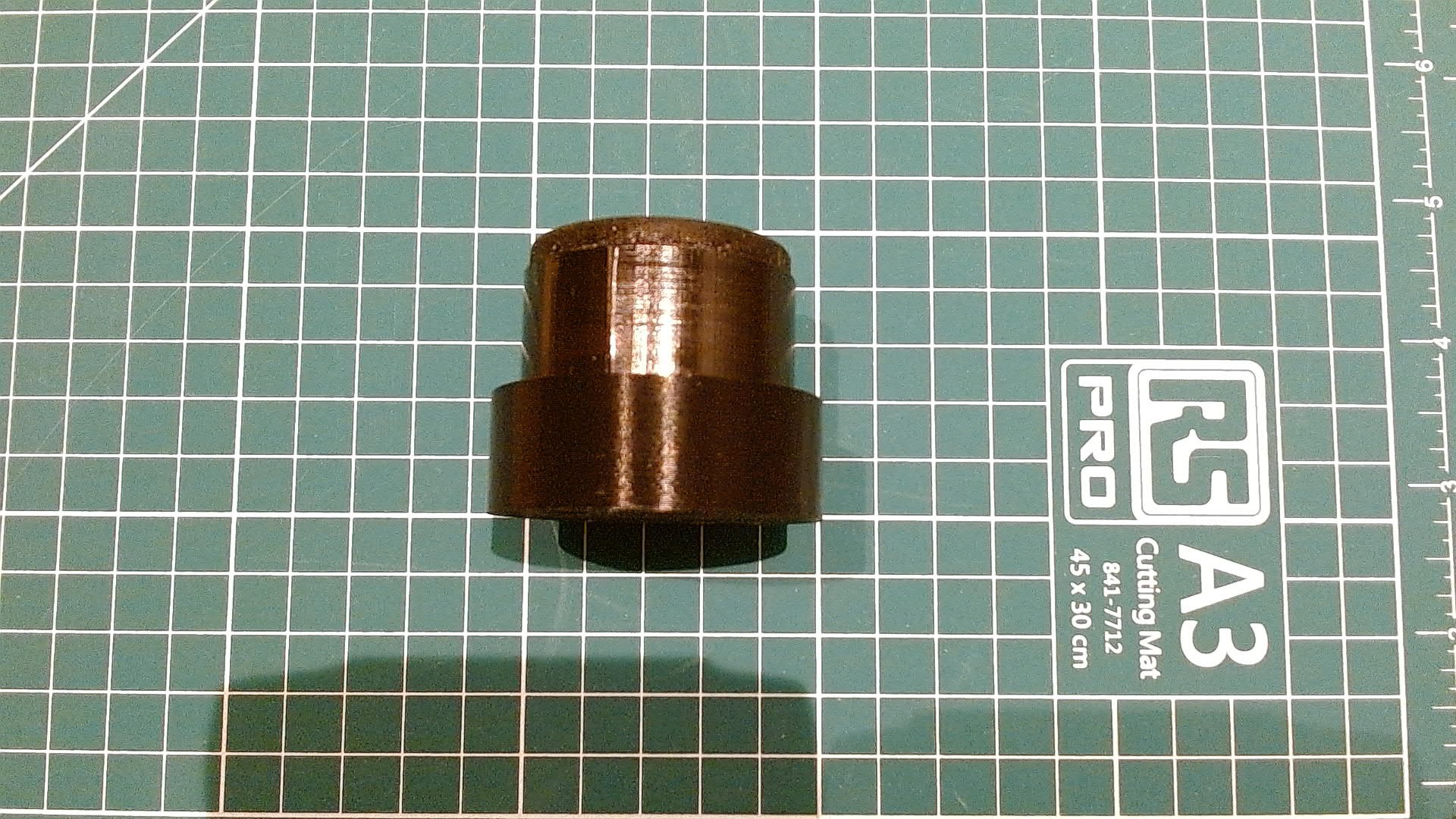} &
		\includegraphics[width=0.41\columnwidth,trim={450 250 550 100},clip]{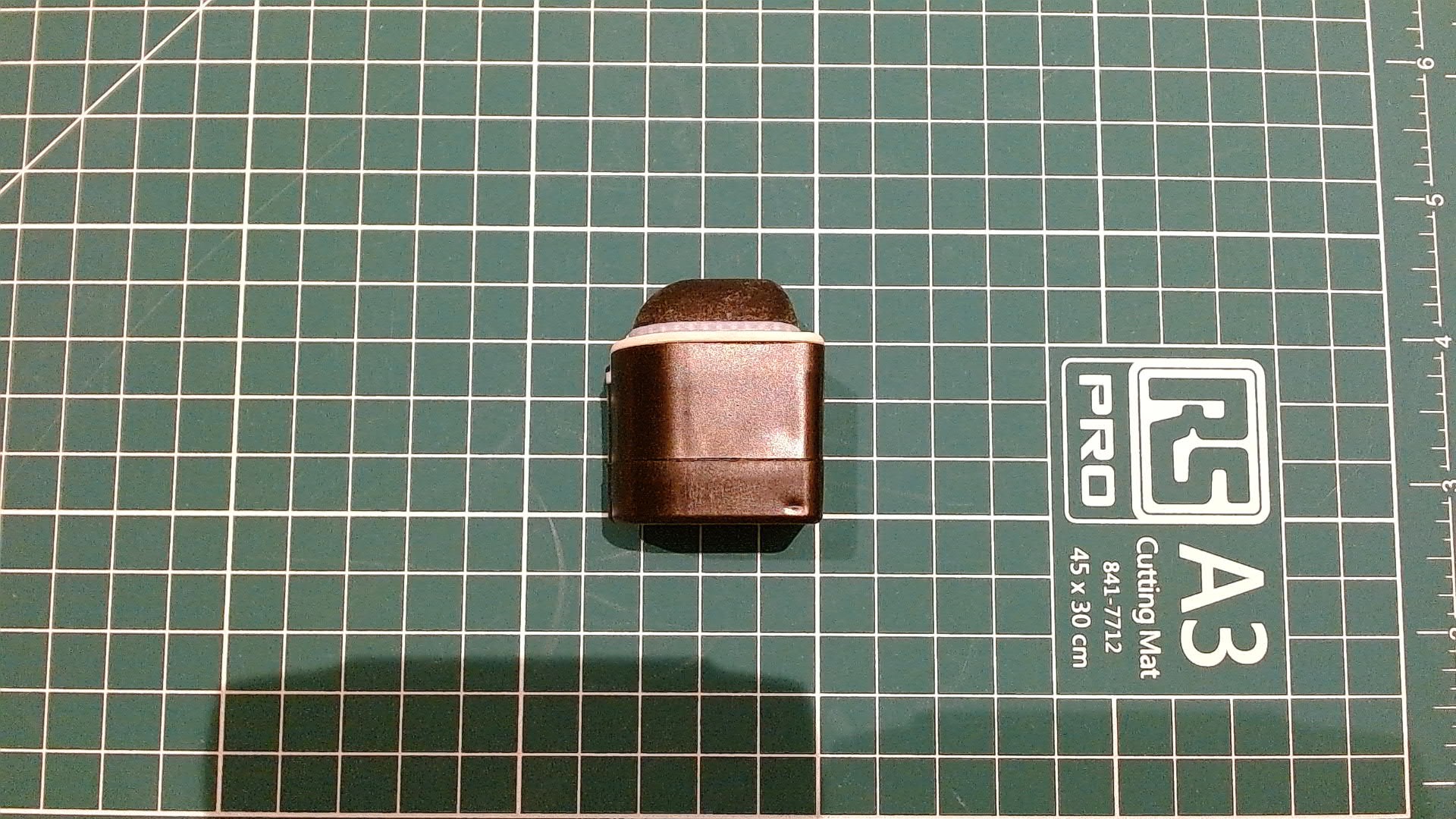} &
		\includegraphics[width=0.41\columnwidth,trim={450 250 550 100},clip]{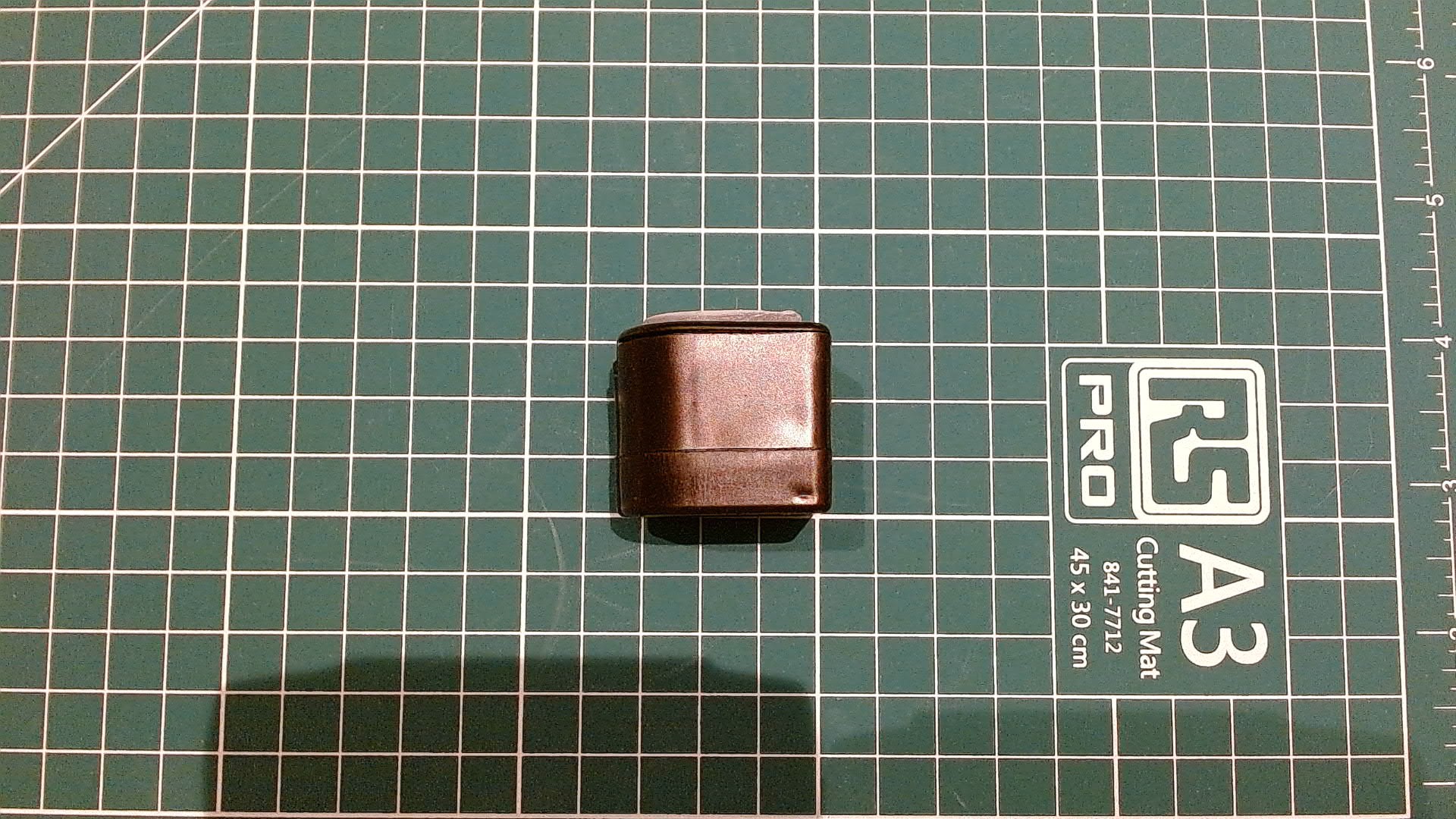} &
		\includegraphics[width=0.41\columnwidth,trim={450 250 550 100},clip]{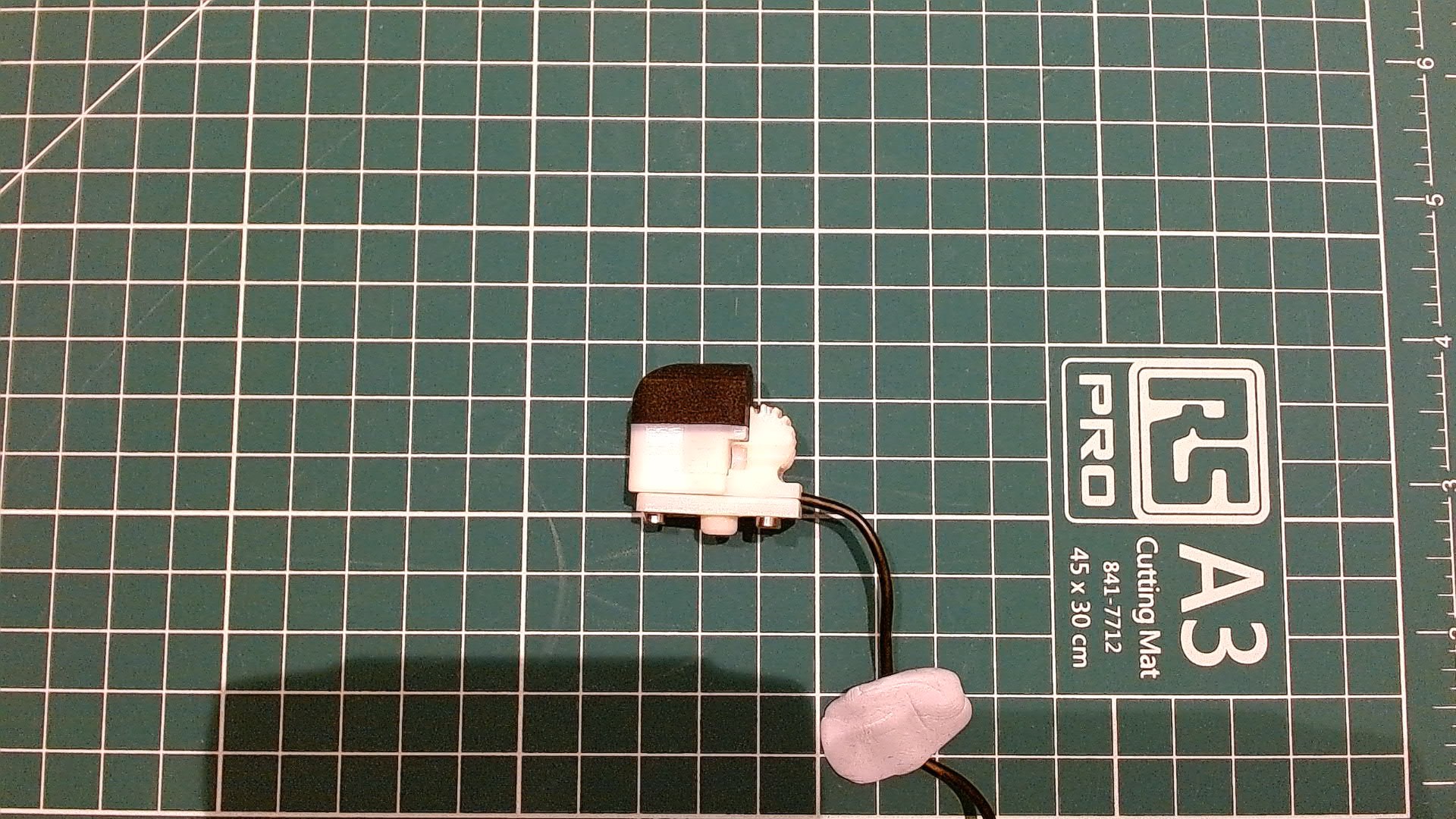} \\
	\end{tabular}
	\caption{Comparative form factors of (a) TacTip (round tip), (b) TacTip (flat tip), (c) DigiTac (curved TacTip skin); (d) DIGIT (flat GelSight skin) and (e) fingertip-sized TacTip~\cite{lepora_towards_2021}. The background grid spacing is 10\,mm.}
	\label{fig:2}
	\vspace{0em}
\end{figure*}

\section{BACKGROUND AND RELATED WORK}

A benefit and constraint of the DIGIT-TACTO-PyTouch ecosystem is the focus on one type of high-resolution optical tactile sensor: the GelSight~\cite{johnson_retrographic_2009-1,yuan_gelsight_2017}. The GelSight images indentation of an elastomer from the surface shading on a reflective coating illuminated by multiple internal light sources. This design concept has diversified into a family of reflection-based optical tactile sensors~\cite{gomes_geltip_2020-2,romero_soft_2020-1,padmanabha_omnitact_2020-1,taylor_gelslim30_2022} including the DIGIT~\cite{lambeta_digit_2020-2}. \textcolor{black}{The first application of deep learning to optical tactile sensing used the GelSight~\cite{yuan_shape-independent_2017}, since leading to many ML models including progress in integrating vision and touch~\cite{luo_editorial_2021}.}

Another type of high-resolution optical tactile sensor uses the morphology of the skin to transduce contact into the motion of markers~\cite{ward-cherrier_tactip_2018-1,lepora_soft_2021-1}. The TacTip is unique in using an array of 3D-printed biomimetic internal pins to amplify the contact due to surface indentation and shear~\cite{lepora_soft_2021-1}. The design is based on human skin that is structured around dermal papillae whose motion is signalled by nearby mechanoreceptors~\cite{chorley_development_2009-1}, with neural signals found recently to match the TacTip output~\cite{pestell_artificial_2022-1}. The tactile images from the TacTip are well-suited to ML models, leading to applications including tactile servo control~\cite{lepora_pose-based_2021,lepora_pixels_2019-1}, pushing manipulation~\cite{lloyd_goal-driven_2022} and sim-to-real transfer of tactile pushing, rolling and servoing~\cite{church_tactile_2021}. A review of these capabilities was published recently~\cite{lepora_soft_2021-1}.

More broadly, there have been many tactile sensor designs utilizing markers~\cite{kamiyama_evaluation_2004}, with recent innovation in embedding markers within and underneath the tactile skin, e.g.~\cite{zhang_deltact_2022,lin_sensing_2019,li_elastomer-based_2019,sferrazza_design_2019}. \textcolor{black}{Some GelSight-type sensors also use markers on the reflective coating~\cite{yuan_gelsight_2017,taylor_gelslim30_2022} to enable sensitivity to shear contact that is not apparent from imaging reflection alone.}

\begin{table}[b!]
	\centering
	\vspace{0em}
	\caption{Comparison of DIGIT, DigiTac and TacTip \textcolor{black}{low-cost high-resolution tactile sensors (each is less than £100 to make)}.}
	\begin{tabular}{c|c|c|c}
	& \textbf{DIGIT} & \textbf{DigiTac} & \textbf{TacTip}  \\
	\hline
	size $l\times w\times h$ [mm] & 36$\times$26$\times$33 & 36$\times$26$\times$39 & 48 dia.$\times$55\\
	weight [g] & 20 & 20 & 65 \\
	sensing field [mm] & 25$\times$19 & 25$\times$19 & 40 dia. \\
	resolution [pix] & 640$\times$480 & 640$\times$480 & 1920$\times$1080\\
	frame rate [per sec]  & 60 & 60 & 120 \\
	biomimetic markers & - & 140 & 331 \\
	\end{tabular}
	\vspace{0em}
	\label{tab:1}
\end{table}

All tactile sensors have their pros and cons. The GelSight and TacTip have a common reliance on imaging an internally-illuminated skin. However, there are differences in their elasticity, robustness, friction and sensitivity to contact features that may make their operation in practise rather different. Our work here aims to bring together knowledge across the field to make it possible to trade-off aspects of these distinct tactile sensors to progress towards tactile-enabled robot dexterity.


\section{\color{black}{Tactile Sensors}}
\label{sec:3}

\subsection{Tactile Sensor Comparison}

\textcolor{black}{The aim of this paper is to compare low-cost high-resolution tactile sensors from the GelSight and TacTip families. For a GelSight-type sensor, we use the DIGIT (Fig.~\ref{fig:2}d), which has a flat sensing surface of 25$\times$19\,mm and VGA camera. We compare this with a domed TacTip (Fig.~\ref{fig:2}a) of 40\,mm diameter and full-HD camera (ELP 1080p USB module). Table~\ref{tab:1} compares the hardware properties of the sensors.}  

\textcolor{black}{However, there is an issue with comparing the sensing capabilities of the DIGIT and TacTip. Any differences in performance may be attributable to the cameras or sensor shapes, rather than the tactile sensing. Therefore, it is important to standardize the hardware and software methods as far as possible to isolate differences due to the sensing mechanism.}

\textcolor{black}{How can two distinct sensors be standardized? Although the TacTip and GelSight-type sensors are superficially similar in both relying on internal cameras, they use different internal illumination. The DIGIT/GelSight has lateral lighting into a thick acrylic window that acts as a light-guide to accentuate shading from the surface indentation. The TacTip has forward-facing lighting situated to both contrast the reflective markers against a black skin and reduce glare on a thin internal window that holds a soft optically-clear gel below the 3D-printed skin.} 

\textcolor{black}{Investigations indicated that the DIGIT could be customized into a TacTip by moving the acrylic window back from the light-source and having a rounded sensing surface. The details of this DIGIT-TacTip customization are described next.}

\subsection{DigiTac Design and Fabrication}
\label{sec:3a}

\afterpage{
\begin{figure}[t]
	\centering
	\begin{tabular}[b]{@{}c@{}}
        \includegraphics[width=\columnwidth]{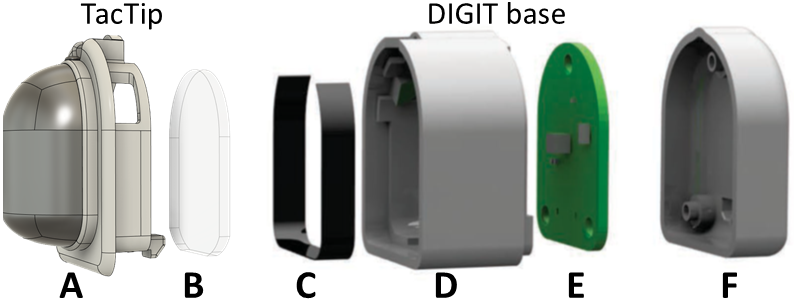} 
    \end{tabular}
	\caption{TacTip sensing surface integrated with DIGIT base. A) 3D-printed TacTip, B) acrylic window, C) lighting PCB, D) plastic housing, E) camera PCB, F) back housing.}
	\label{fig:3}
	\vspace{1.5em}
	\centering
	\begin{tabular}[b]{@{}c@{}}
        \includegraphics[width=\columnwidth]{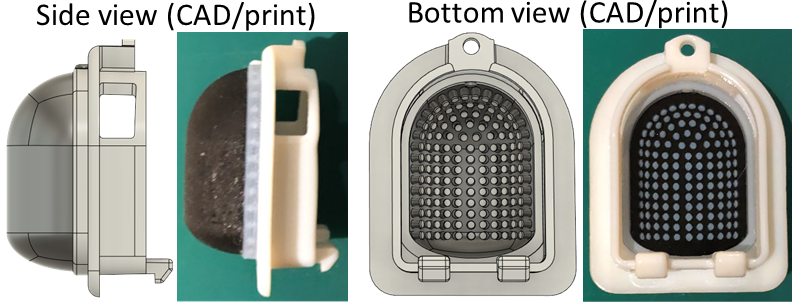} 
    \end{tabular}
	\caption{3D-printable design and fabricated DigiTac sensing surface with press-fit housing.}
	\label{fig:4}
	\vspace{0em}
\end{figure} 
}

The design of the DIGIT optical tactile sensor is based on~\cite{lambeta_digit_2020-2}: (i) being sufficiently compact to fit as an array of end effectors/fingertips on multi-fingered robot hands or arms; (ii) having a sensing surface (gel) that is robust and easily interchanged; and (iii) easing the fabrication process by using off-the-shelf components and a snap-fit assembly. 

In practise, the DIGIT has two distinct assemblies:\\ 
({\bf A1}) The {\bf base unit} comprising the internal lighting PCB, camera PCB and housing (Fig.~\!\ref{fig:3}C-F same as \cite[Fig.~\!2D-G]{lambeta_digit_2020-2});\\ 
({\bf A2}) The {\bf sensing surface} on a thick acrylic window/light-guide and housing that snap fits onto the base. This sensing surface based on the GelSight family of sensors, comprising a painted elastomer, 6\,mm-thick acrylic window and snap-fit holder \cite[\mbox{Fig.~2A-C}]{lambeta_digit_2020-2}. Several elastomers were trialled, including coated markers and a transparent surface~\cite[Fig.~4]{lambeta_digit_2020-2}.

Here we adapt the DIGIT sensing surface into a TacTip sensing surface by:\\
({\bf D1}) The plastic snap-fit holder is redesigned to support a curved compliant skin on the underside of which are 140 2\,mm-papillae tipped by markers (Fig~\ref{fig:4}). The entire piece is fabricated on a multi-material 3D-printer (Stratasys J826), using a rubber polymer (Agilus30, black) for the skin/papillae and acrylic resin (VeroWhite) for the holder/markers.\\
({\bf D2}) The acrylic window is narrowed to 1\,mm thick and moved above the light source to just below the complient skin. The original window acted as a light guide for lateral illumination of the GelSight-type sensing surface. However, the main function of the window for the TacTip is to contain a compliant gel (see D3 below). A thinner window allows more direct marker illumination for a sufficiently curved skin, while also avoiding spurious reflections and saving weight. \\
({\bf D3}) The enclosure between the skin and window is filled with a soft optically-clear gel as in other TacTip sensors~\cite{lepora_soft_2021-1,ward-cherrier_tactip_2018-1}. Here we used a silicone gel (Techsil, Shore A Hardness 15), which was injected through a small (1.5\,mm-dia.) filling hole near the retaining screw on the front of the sensing surface. \textcolor{black}{To stop the gel leaking, the filling hole is then sealed with a small plug, which is also made by 3D-printing.} 

\begin{figure}[t!]
	\centering
	\begin{tabular}[b]{@{}c@{}}
        \includegraphics[width=\columnwidth,trim={110 220 170 180},clip]{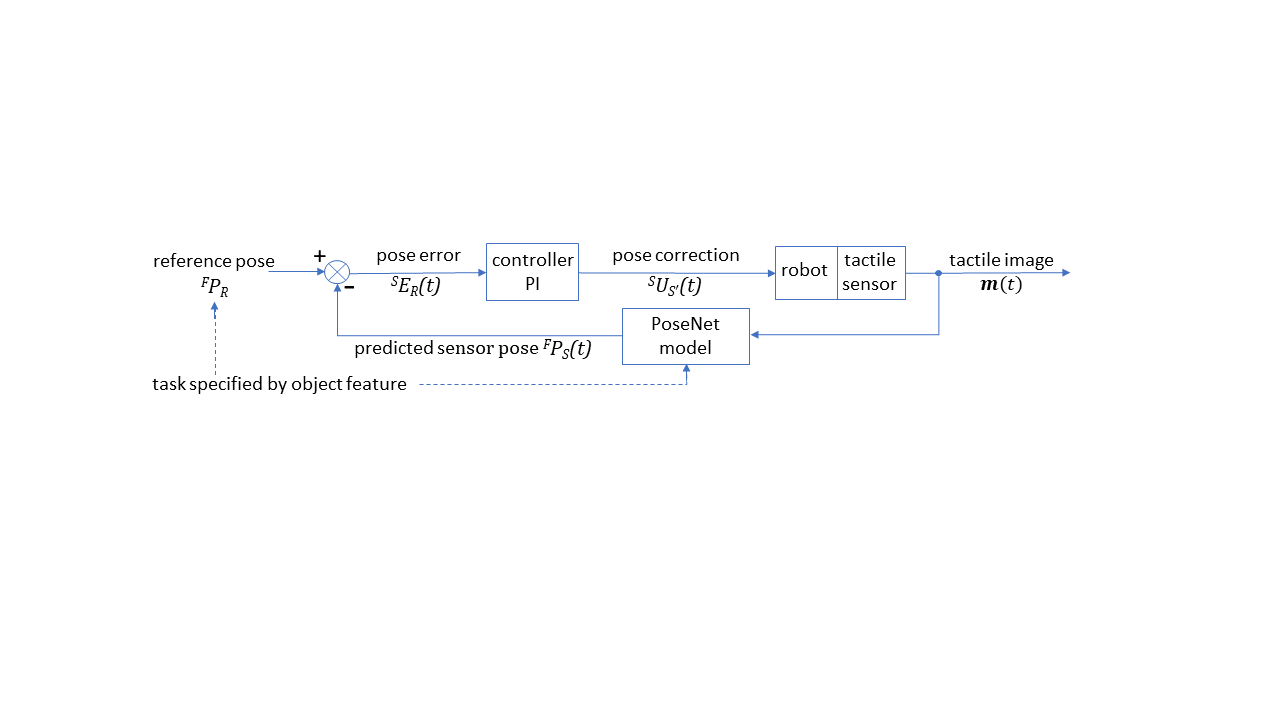} \\
        \includegraphics[width=\columnwidth,trim={205 200 110 150},clip]{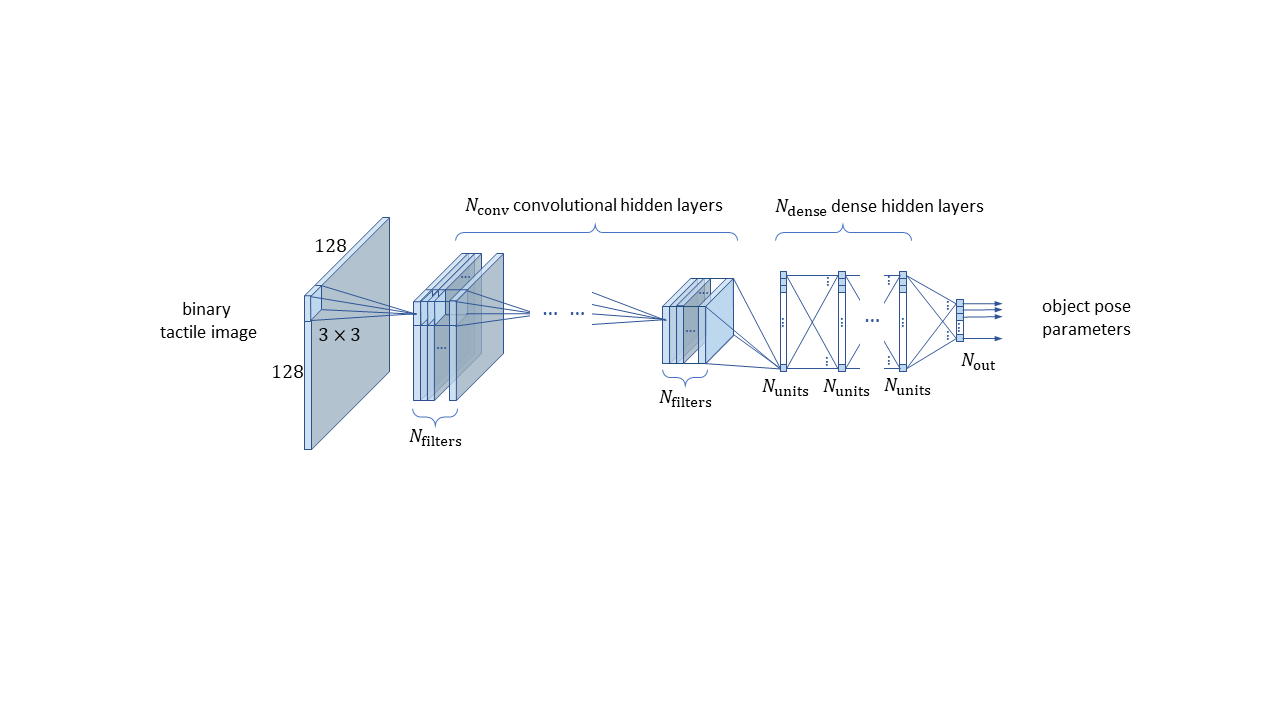} \\
    \end{tabular}
	\caption{Pose-based tactile servo control with PoseNet convolutional neural network model~\cite[Fig. 2]{lepora_pose-based_2021}\cite[Fig. 4]{lepora_optimal_2020-1}.}
	\label{fig:6}
	\vspace{0em}
\end{figure}
\begin{table}[t!]
    \centering
	\caption{Tactile image data collection for sliding contacts: top: pose parameter ranges; bottom: slide ranges to pose. 5000 uniformly-random samples are collected per stimulus.}
	\begin{tabular}{@{}c|cccc@{}}
	 & \multicolumn{4}{c}{\textbf{pose parameter ranges}} \\
	\textbf{stimulus} & $x$-position & $y$-position & $z$-position & angle \\
	\hline
	edge & 0\,mm & [$-5$,$+5$]\,mm & [$-1$,$+1$]\,mm & [$-45^\circ$,$+45^\circ$]\\
	surface & 0\,mm & [$-5$,$-1$]\,mm & 0\,mm & [$-30^\circ$,$+30^\circ$]\\
	\end{tabular}
    \\
    \vspace{1em}	
	\begin{tabular}{@{}c|cccc@{}}
	 & \multicolumn{4}{c}{\textbf{slide parameter ranges}} \\
	\textbf{stimulus} & $x$-position & $y$-position & $z$-position & angle \\
	\hline
	edge & [$-5$,$+5$]\,mm & [$-5$,$+5$]\,mm & 0\,mm & [$-5^\circ$,$+5^\circ$]\\
	surface & [$-5$,$+5$]\,mm & 0\,mm & 0\,mm & [$-5^\circ$,$+5^\circ$]\\
	\end{tabular}
	\label{tab:2}
	\vspace{0em}
\end{table}

The DigiTac sensing surface is assembled in a similar way as other TacTip sensors (see the Soft Robotics Toolkit for a step-by-step guide). The use of multi-material 3D-printing will ease future customization of the design, having already led to a wide range of TacTip form-factors and integrated robotic systems (Fig~\ref{fig:2}; see also \cite[Fig. 4]{lepora_soft_2021-1}). \textcolor{black}{There is a competitive online marketplace for 3D-printing, making it easy to obtain the DigiTac sensing surface from the open-sourced designs.} 


The DigiTac differs from the DIGIT sensor by having a thicker, curved sensing surface (Fig.~2 and Table~I), which we will see later is far softer than the original elastomer. The DIGIT and DigiTac are sufficently small and light to be compatible with some grippers (e.g. the Allegro hand shown in~\cite{lambeta_digit_2020-2}). However, they are too bulky for anthropomorphic robot hands, where a custom TacTip (Fig.~\ref{fig:2}E) is currently the only integrated high-resolution optical tactile sensor.

\section{\color{black}{Experiments}}
\label{sec:3}

\subsection{Task: Tactile Servo Control on a Contour and Surface}
\label{sec:3c}

Here we use servo control around a contour and over a surface to test and compare the various tactile sensors. Tactile servo control is analogous to visual servo control, with tactile data used in a feedback loop to control the robot motion. In pose-based tactile servo control~\cite{lepora_pose-based_2021}, the pose error $^S\!E_R$ between the sensor pose $^F\!P_S$ and a reference pose $^F\!P_R$ (in a local feature frame $F$) drives the feedback loop (Fig.~\ref{fig:6} top). 

For high-resolution optical tactile sensors, the sensor pose $^F\!P_S$ in the feature frame (here of an edge or surface) can be predicted using a suitably-trained `PoseNet' convolutional neural network on a tactile image~\cite{lepora_optimal_2020-1} (Fig.~\ref{fig:6}). For robust pose predictions while interacting with the object, it is necessary to train the model to be insensitive to contact shear motion by sliding the sensor along the object to its labelled pose~\cite{lepora_pixels_2019-1,lepora_optimal_2020-1,lepora_pose-based_2021}. 

Here we use two tactile servo control tasks adapted from the fully 3D-tasks reported in earlier work~\cite{lepora_pose-based_2021} to 2D-tasks implementable on lower-capability desktop robot arms:\\
\noindent {\bf Task 1: Edge-following} is around horizontal flat shapes (here we use circle, square and circular-wave objects described in~\cite{lepora_pose-based_2021}). The PoseNet model is trained on a straight edge of the square block. During training, the sensor pose is varied in its perpendicular distance from the edge and across a range of orientations, using a horizontal sliding motion to move it into position (Table~\ref{tab:2}). Unlabelled variations in contact depth~($z$) are introduced to train the model to be insensitive to sub-mm variations in object height due to imperfect levelling of the robot (which is very difficult to avoid in practise).\\
\noindent {\bf Task 2: Surface-following} is carried out around the walls of the shapes described above with the tactile sensor oriented so the sensing surface is vertical. The PoseNet model is trained on a straight vertical wall of the square block. The labelled pose is varied in distance, normal to the surface, and over a range of surface angles, following a 2D sliding motion tangential to the surface (Table~\ref{tab:2}). A subtlety for servoing over surfaces, is that the tool centre point (TCP) needs adjusting so the $z$-axis of the end-effector frame coincides with the tip of the sensor. 

The PoseNet models are trained on 75\% of 5000 random poses (uniformly distributed), with test accuracies reported on the remaining 25\%. Network hyperparameters are from ref.~\cite[Table~II]{lepora_pose-based_2021} without further tuning, except the input layers were changed to correspond to 160$\times$120 pixel subsampled tactile images for the DIGIT and DigiTac and 128$\times$128 pixels for the TacTip, with the latter two binarized using adaptive thresholding. Networks were trained over 100 epochs ($\sim$30 minutes on a standard PC with 6Gb GPU). 

The tactile sensors are compared using the model performance on the edge and surface test data and the deviation of the controlled trajectories from the known object perimeters. \textcolor{black}{These tests also probe the model generalization for each sensor to curved edges and surfaces not experienced in training.} 

\subsection{Low-cost test platform}
\label{sec:3b}


\begin{figure}[t]
	\centering
	\begin{tabular}[b]{@{}c@{}}
        \includegraphics[width=\columnwidth,trim={10 5 5 5},clip]{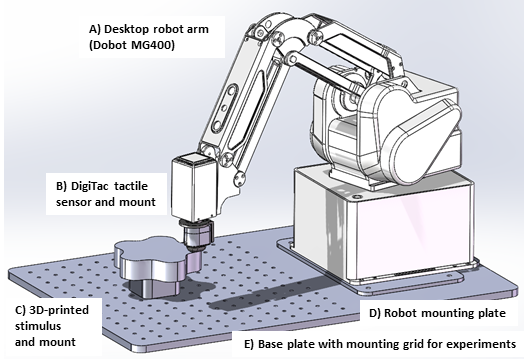} 
    \end{tabular}
	\caption{Schematic of test platform: A) desktop robot arm; B)~tactile sensor on mount; C) 3D-printed stimulus on mount; D) robot mounting plate; E) base plate with mounting grid.}
	\label{fig:5}
	\vspace{-1em}
\end{figure} 

\textcolor{black}{To aid in making the robot test platform accessible to other laboratories, we use a low-cost desktop robot along with laser-cut and 3D-printed accessories for all comparison experiments.}

To give some context, typically the hardware and software infrastructure required for researching robot touch has been:\\
\noindent ({\bf I1}) Robot manipulators, such as ABB and Universal Robotics industrial robot arms. These are highly capable and accurate (e.g. ABB-IRB120 repeatability $\pm0.01$\,mm), but are large, expensive items of equipment that must be installed and operated safely in a laboratory or industrial setting.\\
\noindent ({\bf I2}) Software libraries for capturing tactile data, processing with ML models and controlling the robots in real-time. Different laboratories have various solutions. In \textcolor{black}{Bristol Robotics Laboratory}, we have developed Python libraries, VSP \textcolor{black}{(Visual Stream Processor)} and CRI \textcolor{black}{(Common Robot Interface)} for asynchronous tactile image capture and interfacing with proprietary robot APIs, available on GitHub.

Here we introduce a test platform to ease the setup and repeatability of tactile experiments using a low-cost safe alternative to industrial robot arms (Fig.~\ref{fig:5}):\\
\noindent ({\bf P1}) Desktop robot arm, for which we use a Dobot~MG400 4-axis arm designed for affordable automation. The base and control unit has footprint 190\,mm$\times$190\,mm, payload 750\,g, maximum reach 440\,mm and repeatability $\pm0.05\,$mm. As we describe later, the accuracy of tactile models trained using this arm is similar to larger industrial robot arms. The main constraint is that only the $(x,y,z)$-position and rotation around the $z$-axis of the end effector are actuated. \\
\noindent ({\bf P2}) Base plate with a mounting plate for the desktop robot and a grid of holes for mounting stimuli in precise locations relative to the robot. The base plate was laser cut from an acrylic sheet of size 600$\times$400$\times$10\,mm, with 25\,mm grid-spacing. The robot mounting plate was also laser cut from 10\,mm-thick acrylic then screwed to the base plate.\\ 
\noindent ({\bf P3}) Normal and right-angle mounts for the DIGIT and TacTip sensors so they can be used as end effectors. These were 3D-printed from designs adapted from an end-flange supplied with the MG400. A choice of mounts allows the tactile sensing surface to be oriented in a horizontal or vertical plane, as the end effector can rotate only around a vertical axis.\\ 
\noindent ({\bf P4}) Test stimuli with mounts to attach to the base plate. Here we use circular, square and circular-wave shapes of $\sim$100\,mm diameter with 30\,mm walls, 3D-printed in ABS.


\section{RESULTS}
\label{sec:4}

\begin{figure}[t!]
	\centering
	\begin{tabular}[b]{@{}c@{\hspace{2pt}}c@{\hspace{2pt}}c@{}}
		{\bf (a) DIGIT} & {\bf (b) DigiTac} & {\bf (c) TacTip}\\
		\includegraphics[width=0.33\columnwidth,trim={500 350 450 0},clip]{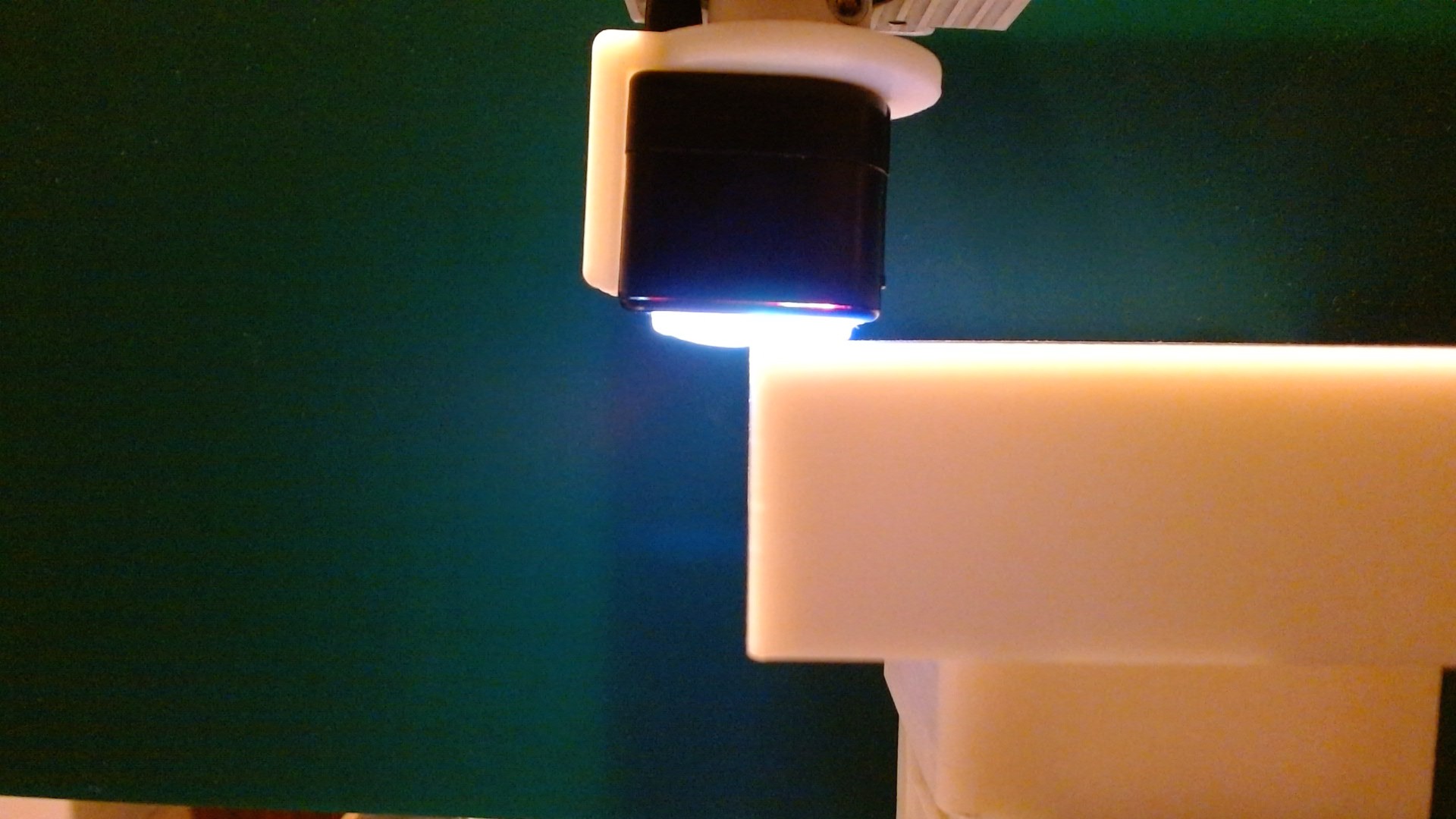} &
		\includegraphics[width=0.33\columnwidth,trim={500 350 450 0},clip]{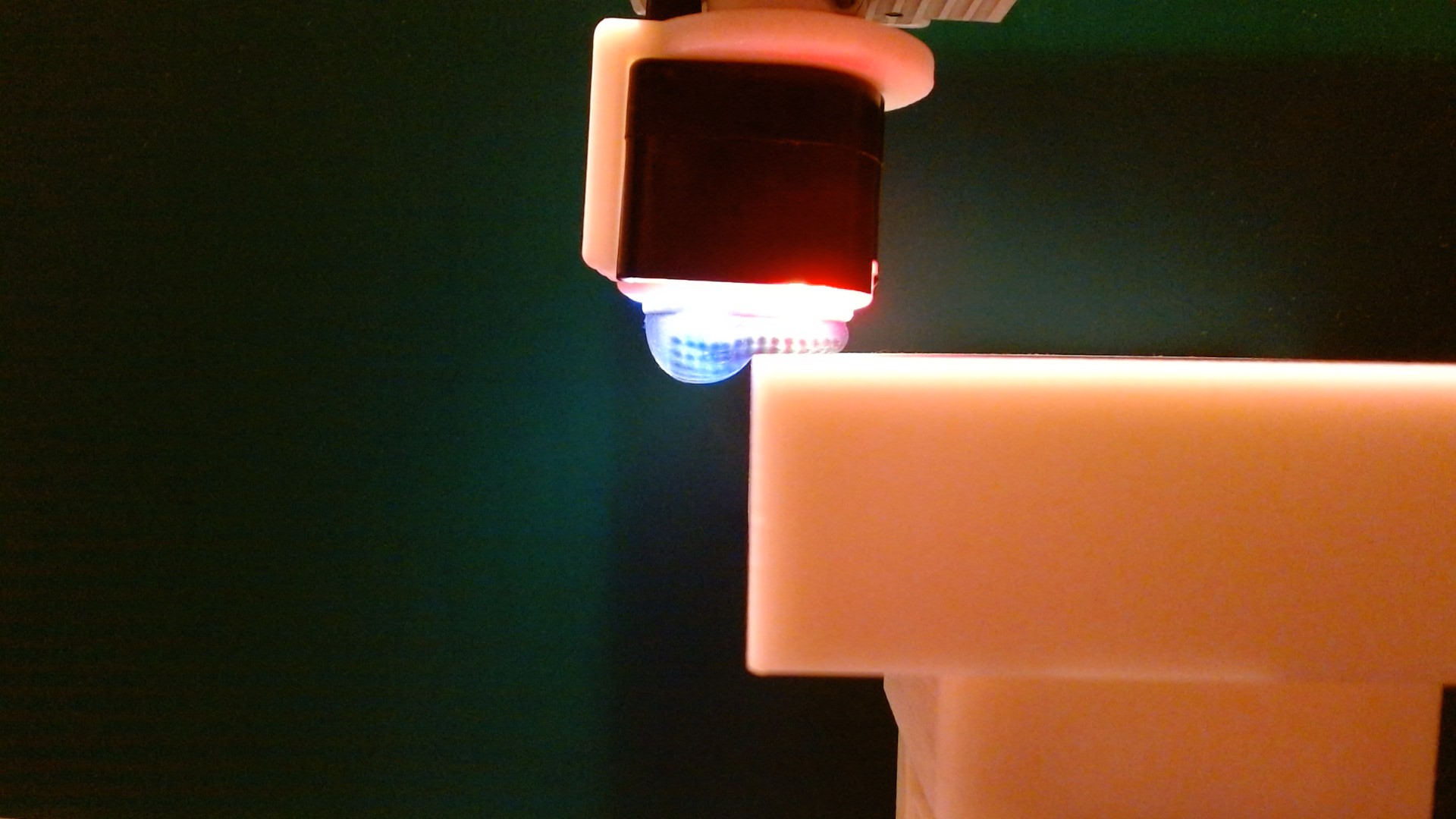} &
    	\includegraphics[width=0.33\columnwidth,trim={500 330 450 20},clip]{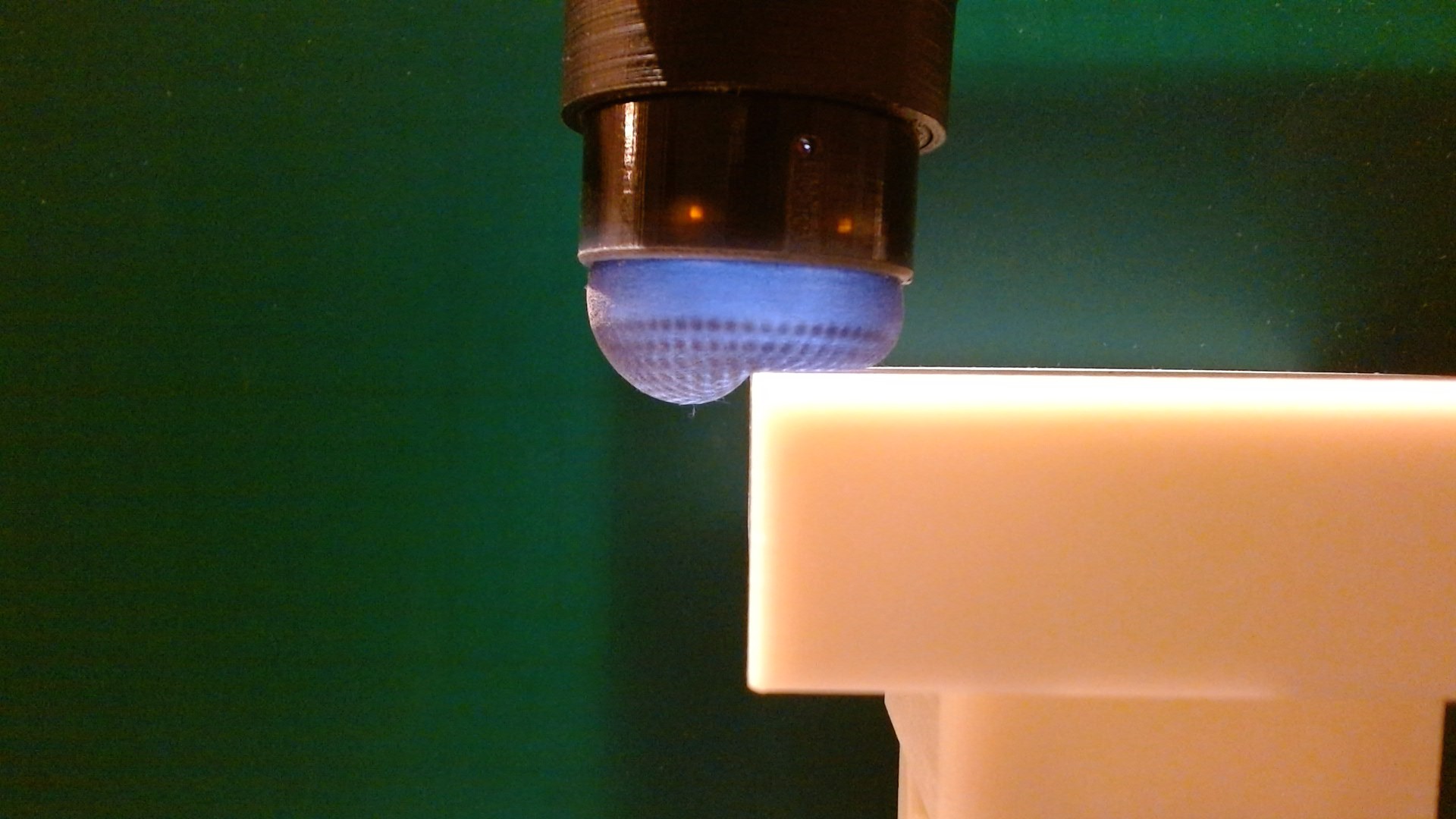} \\
    	\includegraphics[width=0.33\columnwidth,trim={0 0 0 0},clip]{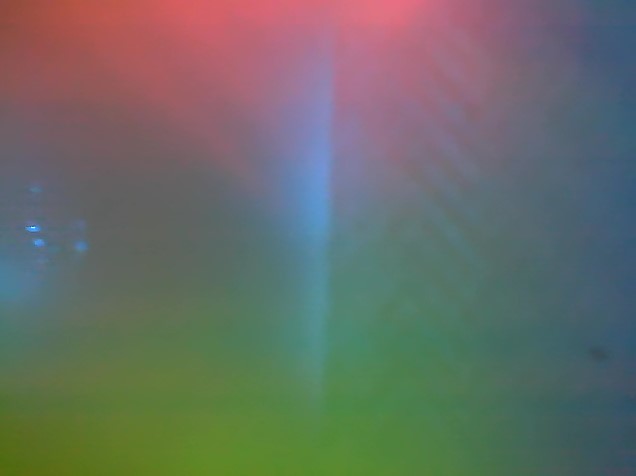} &
		\includegraphics[width=0.33\columnwidth,trim={0 0 0 0},clip]{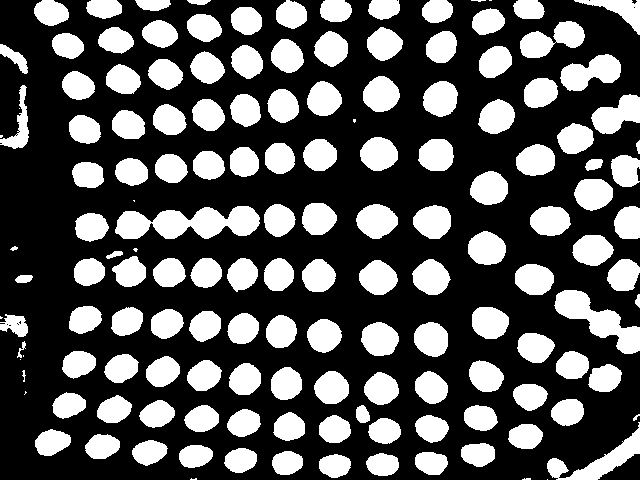} &
    	\includegraphics[width=0.25\columnwidth,trim={0 0 0 0},angle=90,clip]{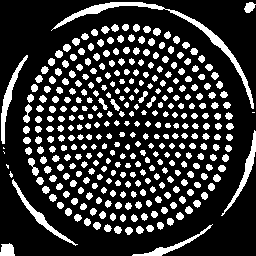} \\
		{\bf (d) RGB image} & {\bf (e) BW image} & {\bf (f) BW image}\\
	\end{tabular}
	\caption{Top: views of the DIGIT, DigiTac and TacTip (round) pressing on an edge. Bottom: corresponding tactile images.}
	\label{fig:7}
	\vspace{-0em}
\end{figure}

\subsection{Appraisal of DIGIT, DigiTac and TacTip}
\label{sec:4a}

For an initial qualitative comparison of the three optical tactile sensors, we pressed them against a 3D-printed straight edge and captured the tactile images (Fig.~\ref{fig:7}). 

From the external appearance of the deformation, the Digi-Tac and TacTip (Figs~\ref{fig:7}b,c) are much softer than the DIGIT (Fig.~\ref{fig:7}a), with a large compression of several millimeters over the contact region and a corresponding bulging of part of the sensing surface not in contact. In contrast, the DIGIT sensing surface is fairly stiff with sub-mm deformation. The maximum compression of the DigiTac or TacTip is about 6-8\,mm before damage will occur and for the DIGIT about 1-2\,mm. 

Another difference is that the DIGIT sensing surface has higher friction than the DigiTac. It is fairly easy to slide a DigiTac or TacTip over a 3D-printed flat surface with slight texture, whereas the DIGIT tends to stick then slip.

\afterpage{
    \begin{figure}[t!]
    	\centering
    	\begin{tabular}[b]{@{}c@{\hspace{10pt}}c@{}}
    		\multicolumn{2}{c}{\bf DigiTac: Edge pose -- displacement and contact angle}\\
    		\includegraphics[width=0.7\columnwidth,trim={20 0 30 10},clip]{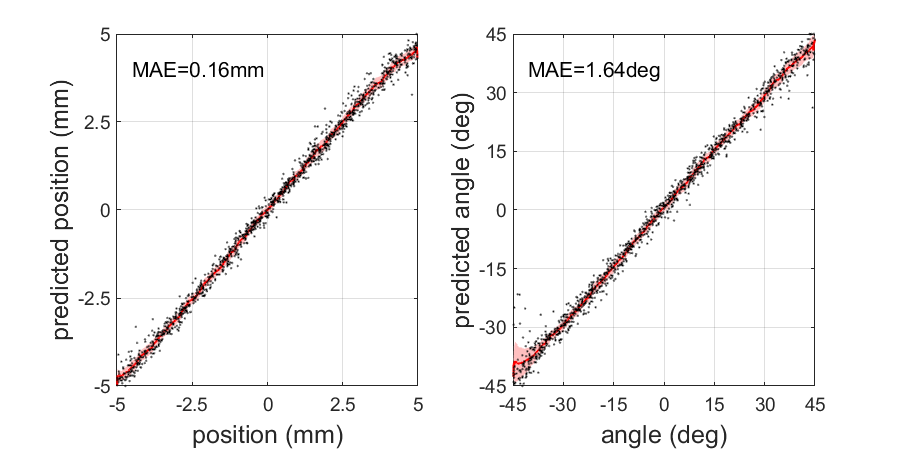} & 		
    		\includegraphics[width=0.28\columnwidth,trim={500 200 450 0},clip]{fig4d_alt.jpg} \\
    		\multicolumn{2}{c}{\bf DigiTac: Surface pose -- depth and contact angle}\\
    		\includegraphics[width=0.7\columnwidth,trim={20 0 30 10},clip]{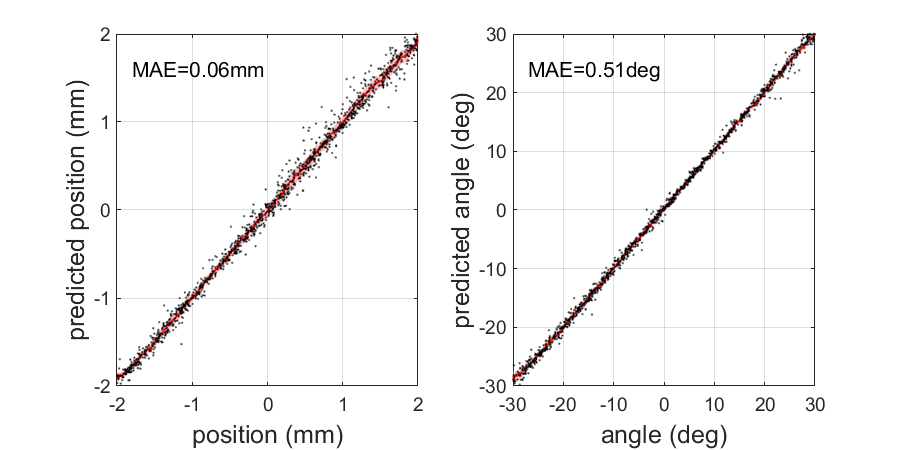} & 	
    		\includegraphics[width=0.28\columnwidth,trim={325 50 625 150},clip]{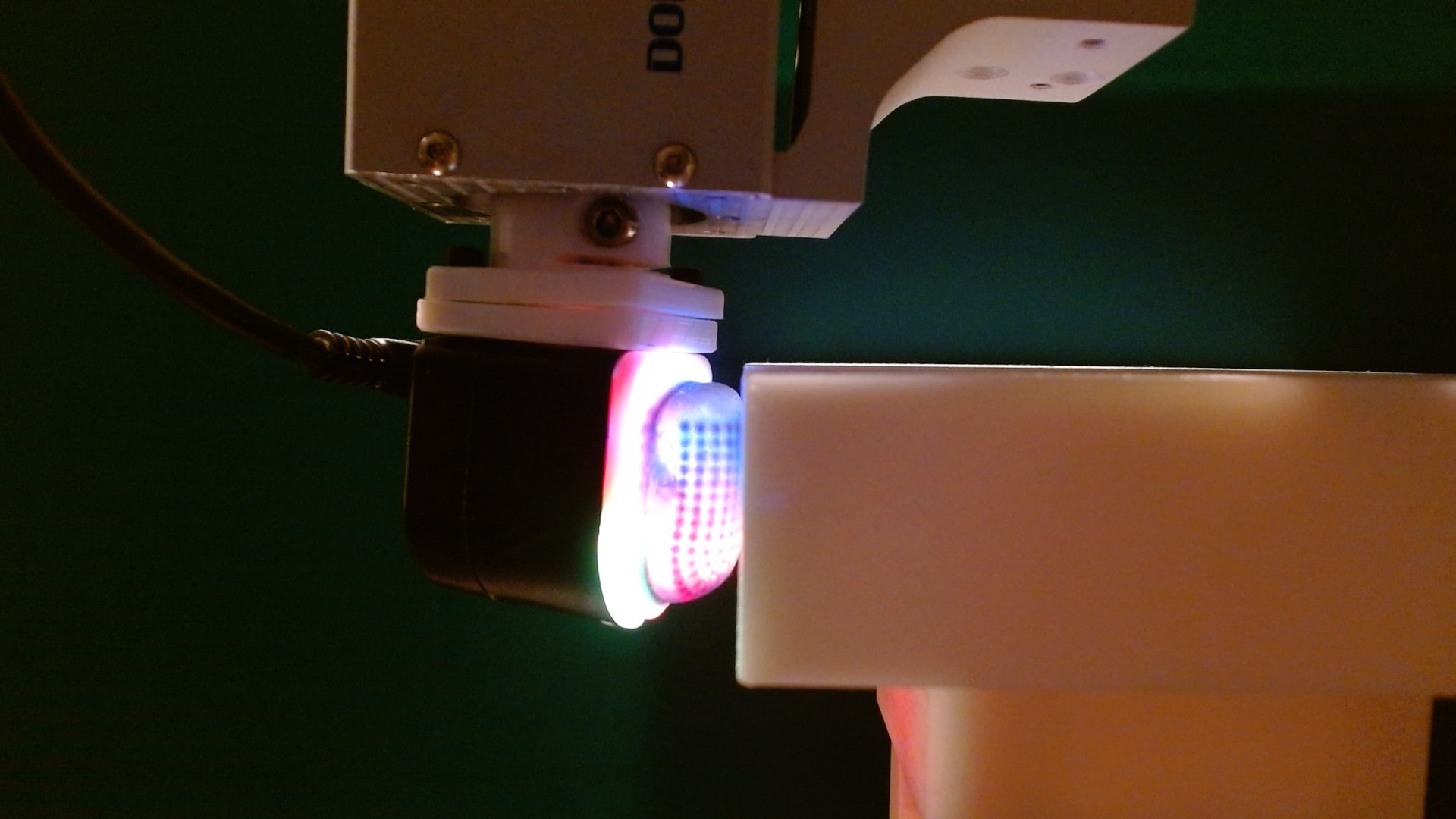} \\
    	\end{tabular}
    	\vspace{-1em}
    	\caption{Pose prediction with the DigiTac. Top: edge displacement and angle. Bottom: surface depth and angle. Predictions are under random unknown sliding contacts (Table~\ref{tab:2}).}
    	\vspace{0em}
    	\label{fig:8}
    \end{figure}
    \vspace{0em}
    \begin{table}[h]
    	\caption{Accuracy (MAE) of pose estimation for the DIGIT, DigiTac and TacTip on the straight edge and flat surface. Predictions are under random sliding contacts (Table~\ref{tab:2}).}
    	\vspace{0em}
    	\centering
    	\begin{tabular}{c|ccc}
    	\textbf{sensor} & \textbf{stimulus} & \textbf{position $y$} & \textbf{angle} \\
    	& & \textbf{MAE / range} & \textbf{MAE / range} \\
    	\hline
    	DIGIT & edge & 0.19\,mm / 10\,mm & 1.83$^\circ$ / 90$^\circ$\\
    	& surface & \textcolor{black}{0.22\,mm  / 1\,mm} & \textcolor{black}{2.47$^\circ$ / 60$^\circ$} \\
    	\hline	DigiTac & edge & 0.16\,mm / 10\,mm & 1.64$^\circ$ / 90$^\circ$\\
    	 & surface & 0.06\,mm /\ \ 4\,mm & 0.51$^\circ$ / 60$^\circ$ \\
    	\hline
    	TacTip & edge & 0.16\,mm / 10\,mm & 1.66$^\circ$ / 90$^\circ$ \\
    	 & surface & 0.11\,mm /\ \ 4\,mm & 0.73$^\circ$ / 60$^\circ$ \\
    	\end{tabular}
    	\label{tab:3}
    	\vspace{0em}
    \end{table}
}

From the tactile images, the edge is clearly visible with the DIGIT as a shaded line at the point of contact~(Fig.~\ref{fig:7}d). For the DigiTac and TacTip, the edge can be located by where the markers are more widely spaced (Figs~\ref{fig:7}e,f). Thus, it is more difficult to see edge location by eye; however, we will see later that the edge can be precisely localized by using a convolutional neural network on the entire tactile image.

The DIGIT is sensitive enough for the texture of the 3D-print to be just visible (Fig~\ref{fig:7}d, right of edge). This information is absent from static touch with the DigiTac/TacTip. In this respect, the DIGIT is superior to the human sense of touch, which does not have the sensitivity to feel that the 3D-print is textured by statically holding a fingertip on the surface.

\subsection{Sensor Comparison using Edge Following}
\label{sec:4b}

Next, we compare the tactile sensors on predicting edge position and angle during sliding contacts, then use this pose model for tactile servo control around the edges of three test objects: a circular disk, square block and circular-wave. 

For offline edge-pose prediction on the test data, both the DigiTac and TacTip are similarly accurate with edge position error 0.16\,mm and angle error $\sim$1.6$^\circ$ (Table~\ref{tab:3}). These errors are $\sim$1.5\% of the ranges (10\,mm and 90$^\circ$), with little scatter when the predictions are plotted against ground truths (Fig.~\ref{fig:8}). It appears the smaller size of the DigiTac and fewer markers (Table~\ref{tab:1}) does not affect its capability for predicting edge pose. 

In comparison, the DIGIT has similar edge position error (0.19\,mm vs 0.16\,mm) and angle (1.82$^\circ$ vs $\sim$1.6$^\circ$) compared with the DigiTac. To isolate the effects of sensor surface material and geometry from subsequent processing, we used the same neural network architectures and hyperparameters with binary tactile images for the DigiTac and shaded images for the DIGIT. While these results could change with a different choices of neural networks, all accuracies are sub-mm and around 1-2 degrees, which are appropriate for the contour-following experiments reported below. 

\begin{figure}[t!]
	\centering
    \begin{tabular}[b]{@{}c@{\hspace{6pt}}c@{}}
    	\textbf{\footnotesize{\ \ DIGIT: Edge Following}} &  \\
	    \includegraphics[width=0.49\columnwidth,trim={28 100 36 35},clip]{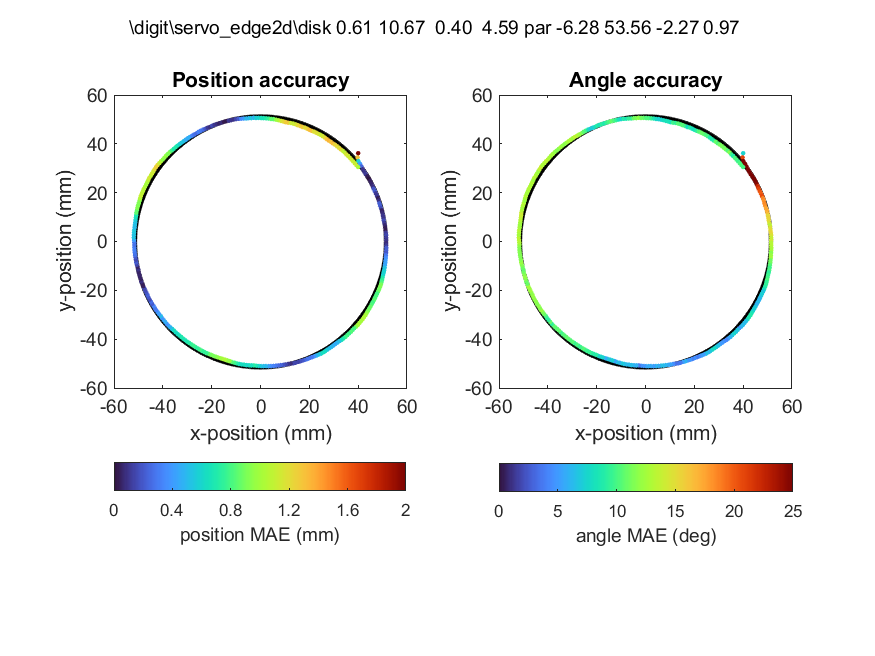} & 
	    \includegraphics[width=0.35\columnwidth,trim={0 200 0 0},clip]{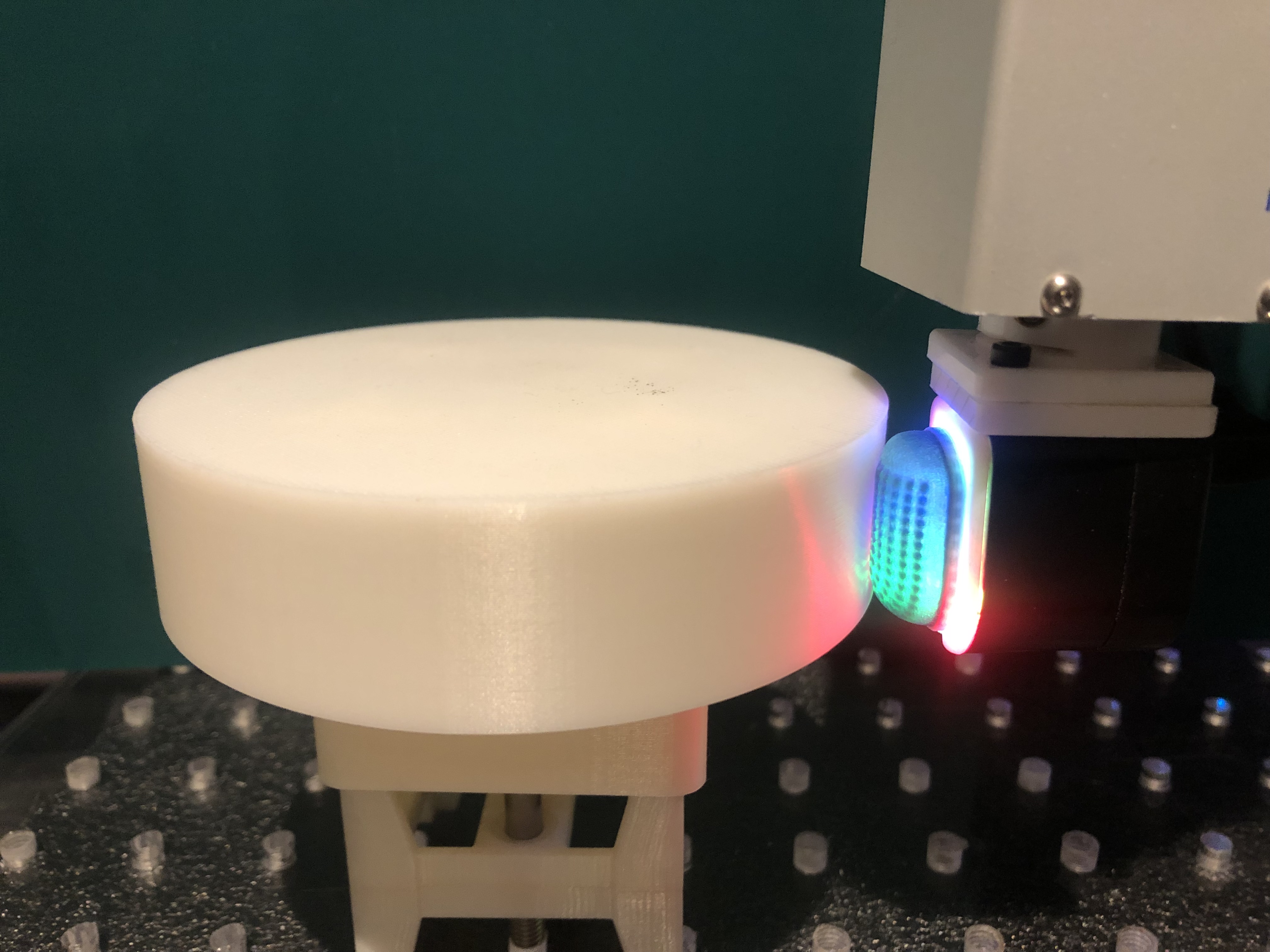}\\
		 \includegraphics[width=0.49\columnwidth,trim={28 100 36 35},clip]{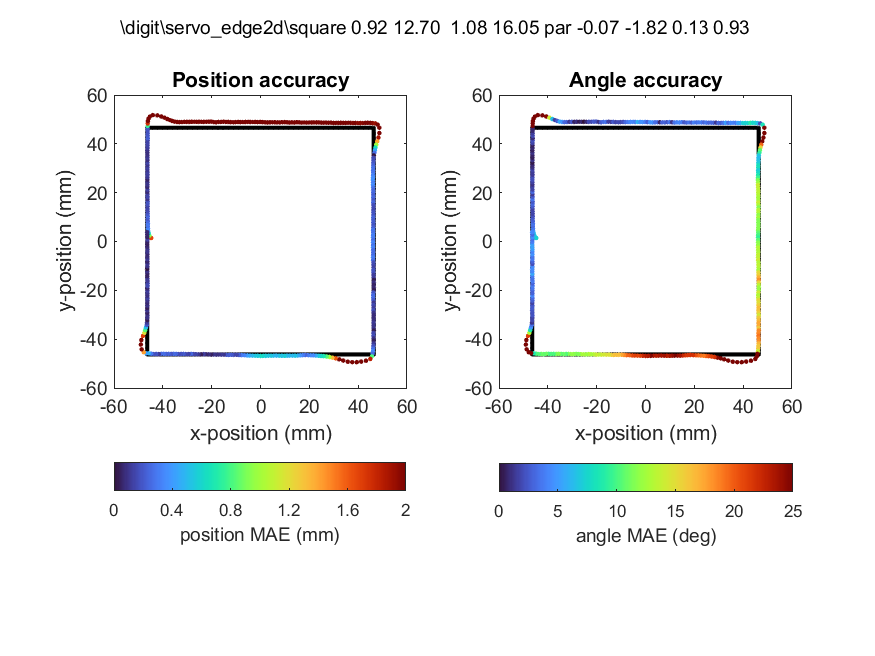} & 
	    \includegraphics[width=0.35\columnwidth,trim={200 400 200 200},clip]{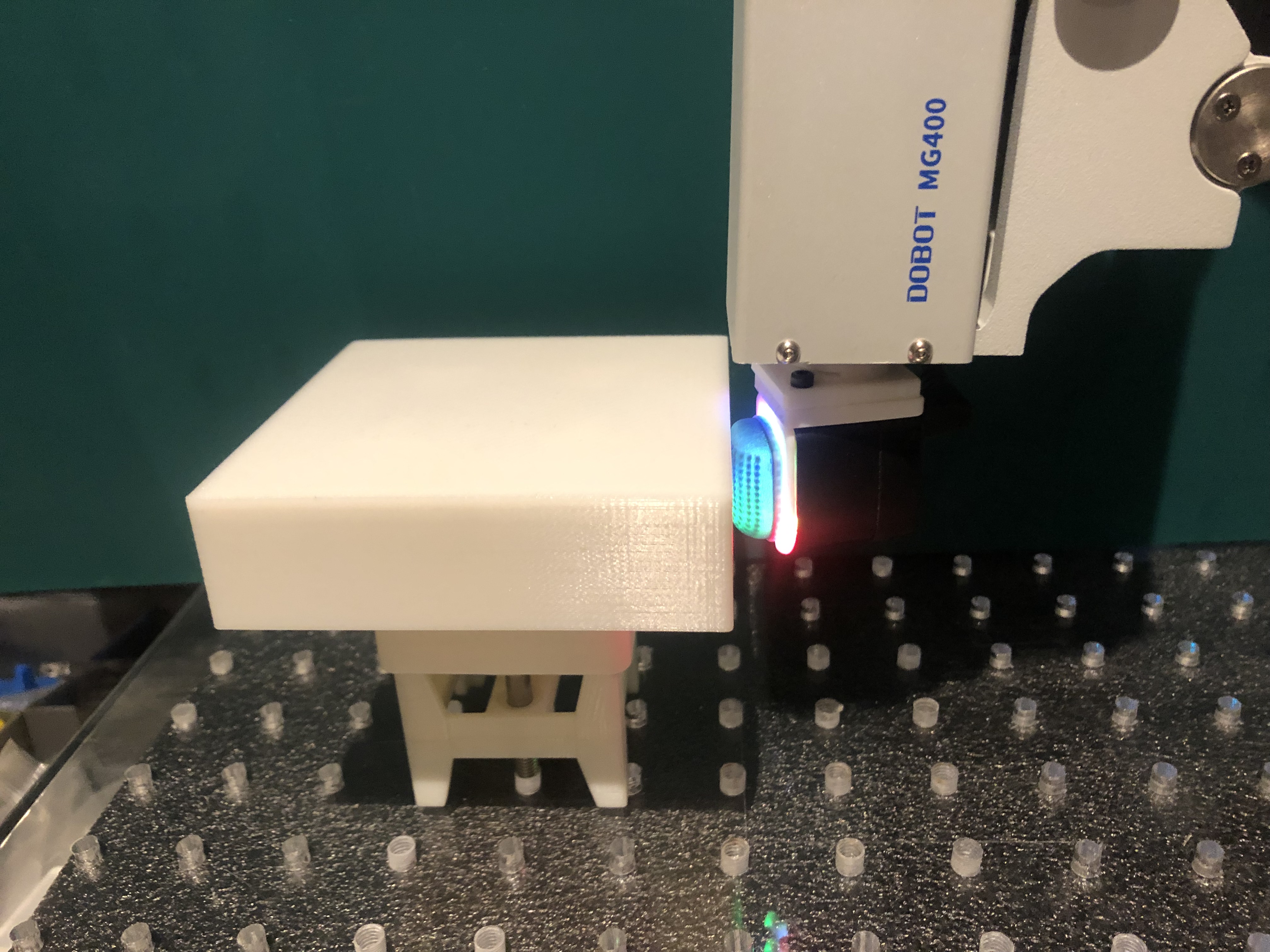}\\
		 \includegraphics[width=0.49\columnwidth,trim={28 100 36 35},clip]{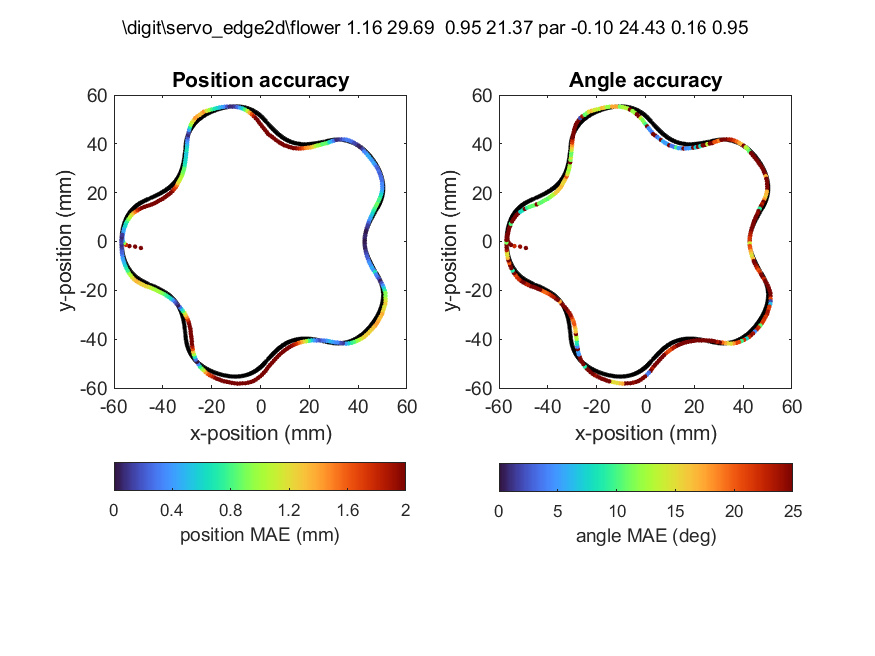} & 
	    \includegraphics[width=0.35\columnwidth,trim={200 400 200 200},clip]{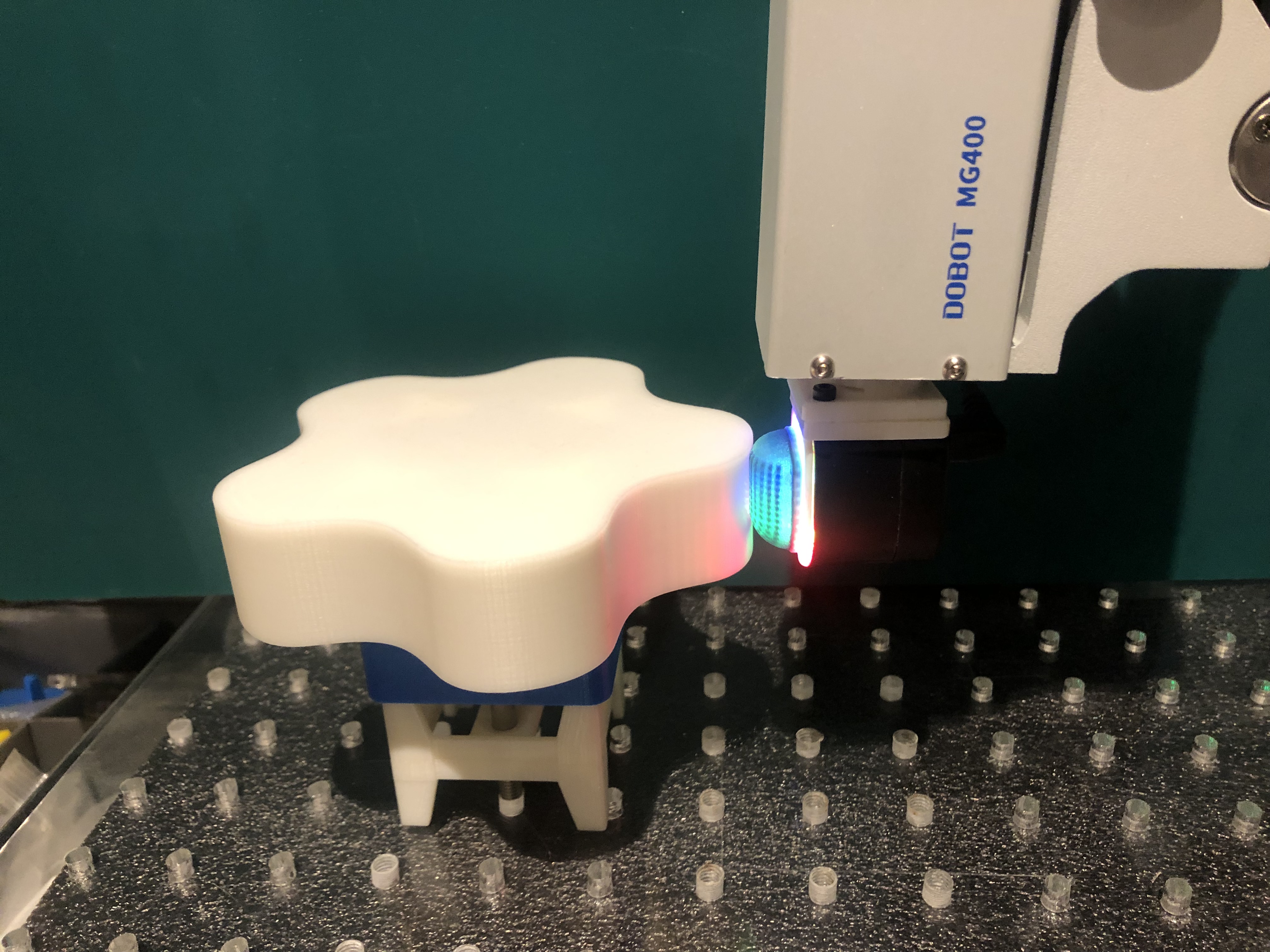}\\
        \textbf{\footnotesize{\ \ DigiTac: Edge Following}} & \textbf{\footnotesize{DigiTac: Surface Following}} \\
	    \includegraphics[width=0.49\columnwidth,trim={28 100 36 35},clip]{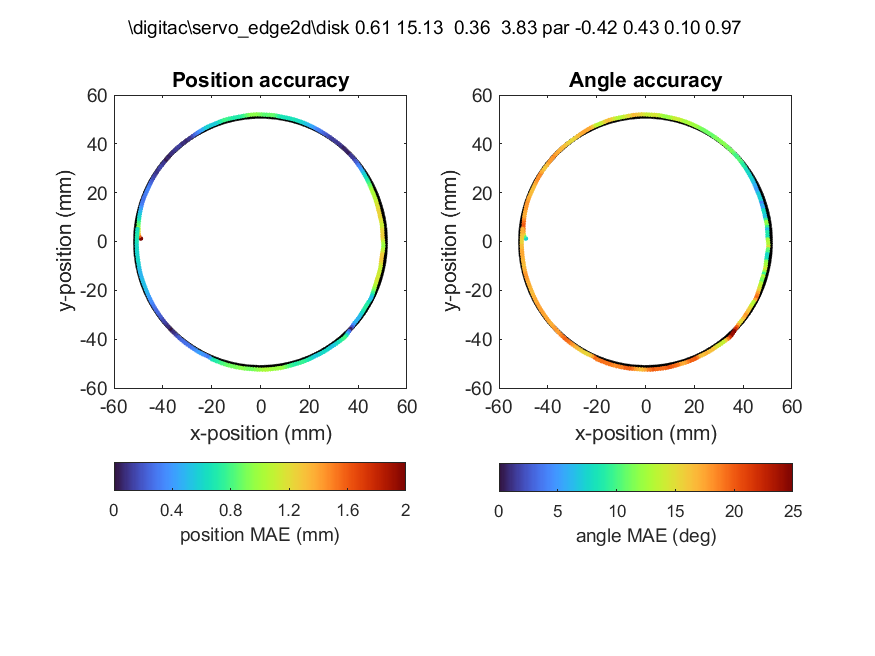} & 		
		\includegraphics[width=0.49\columnwidth,trim={28 100 36 30},clip]{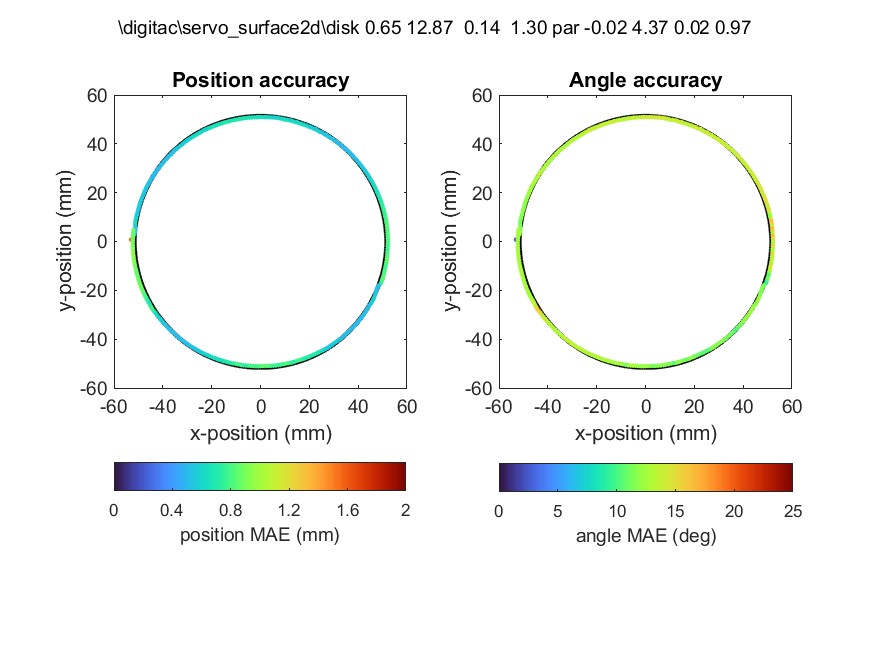} \\
        \includegraphics[width=0.49\columnwidth,trim={28 100 36 30},clip]{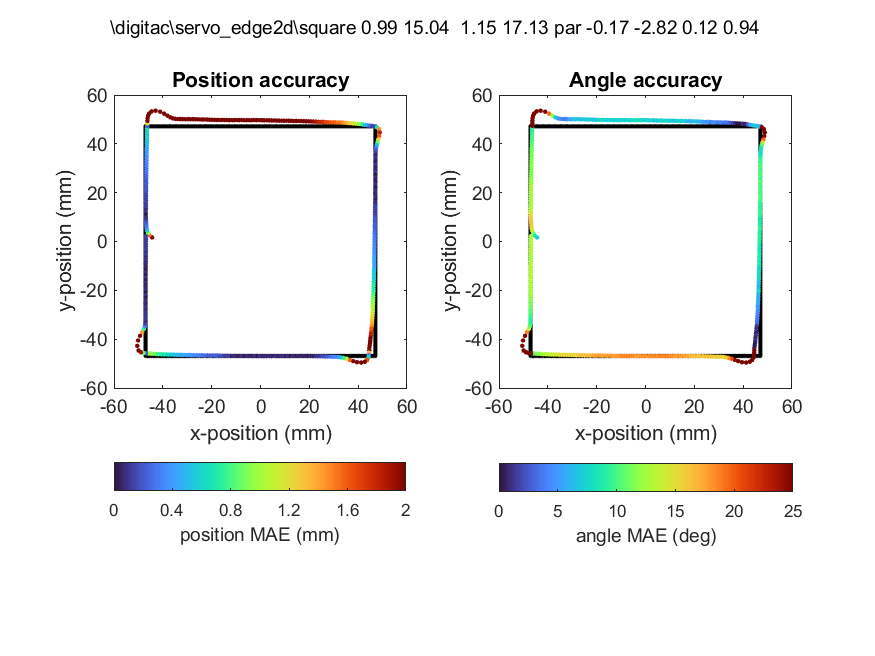} & 		
		\includegraphics[width=0.49\columnwidth,trim={28 100 36 30},clip]{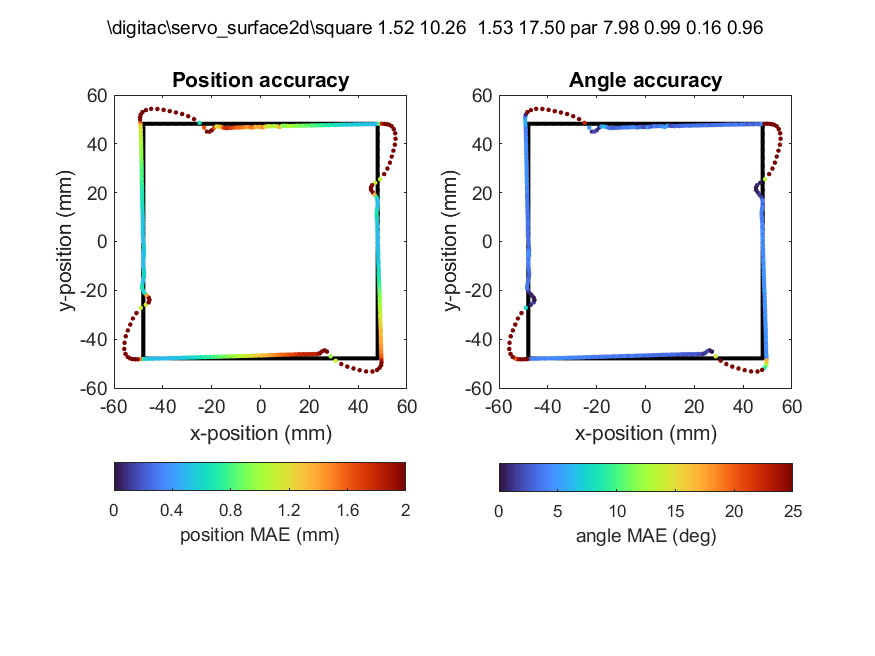} \\
		\includegraphics[width=0.49\columnwidth,trim={28 100 36 30},clip]{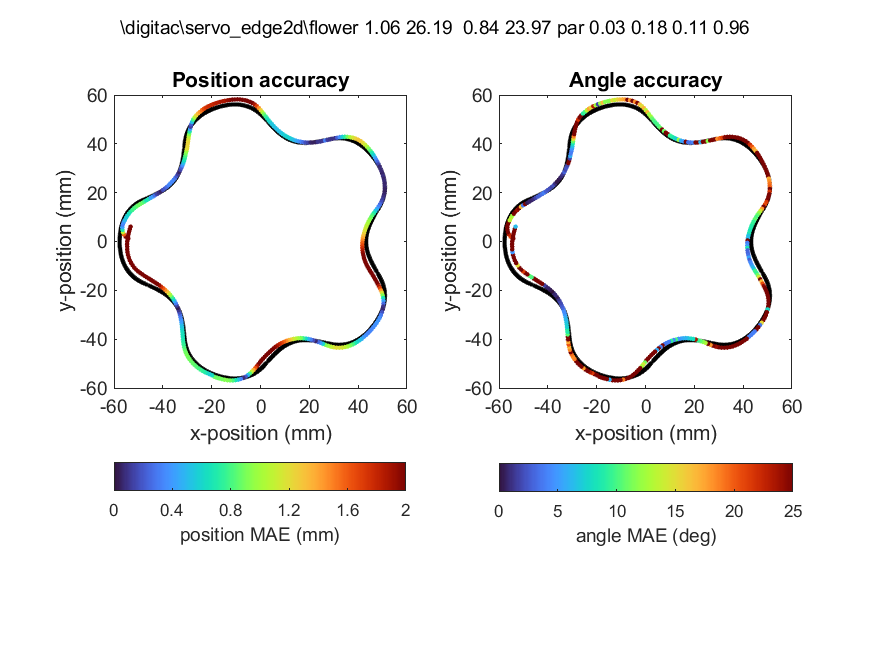} & 		
		\includegraphics[width=0.49\columnwidth,trim={28 100 36 30},clip]{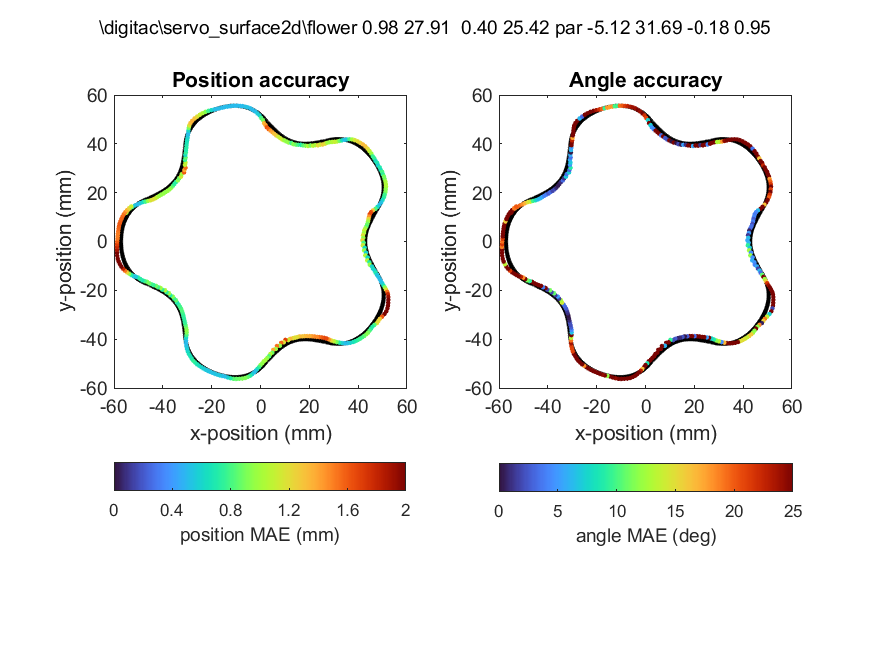} \\
		\textbf{\footnotesize{TacTip: Edge following}} & \textbf{\footnotesize{TacTip: Surface following}} \\
		\includegraphics[width=0.49\columnwidth,trim={28 100 36 30},clip]{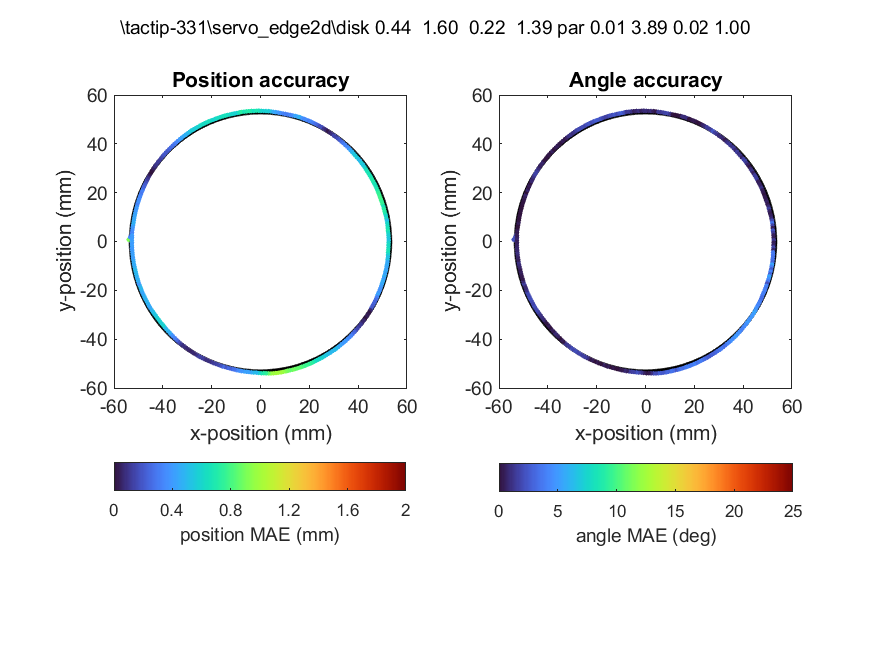} & 		
		\includegraphics[width=0.49\columnwidth,trim={28 100 36 35},clip]{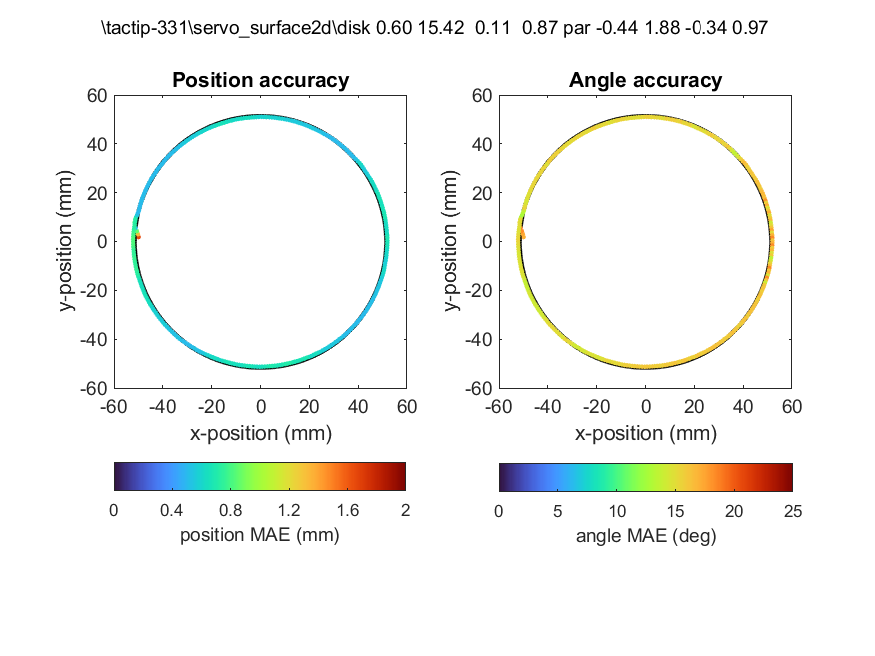} \\		
	    \includegraphics[width=0.49\columnwidth,trim={28 100 36 35},clip]{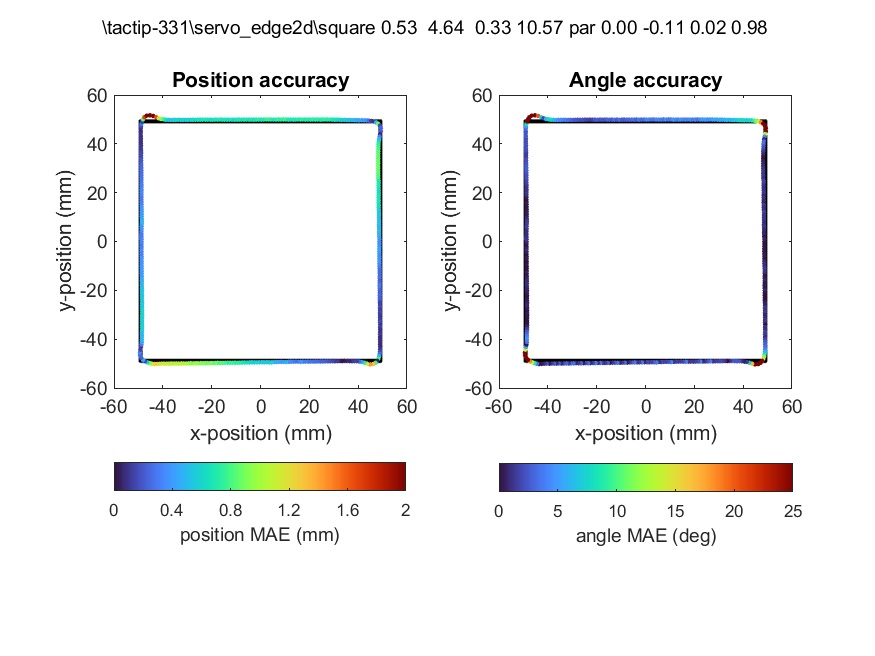} & 		
		\includegraphics[width=0.49\columnwidth,trim={28 100 36 35},clip]{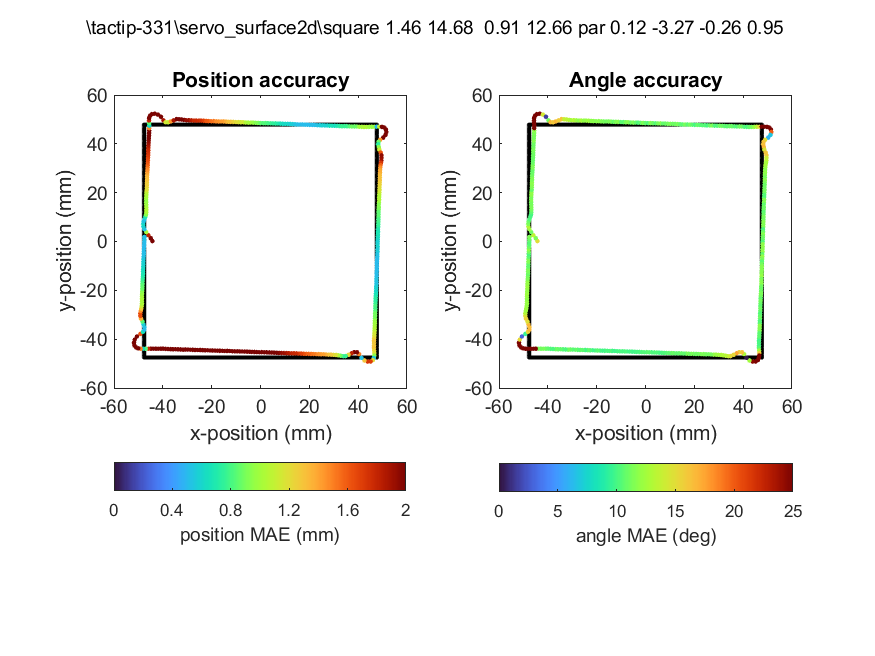} \\
	    \includegraphics[width=0.49\columnwidth,trim={28 50 36 35},clip]{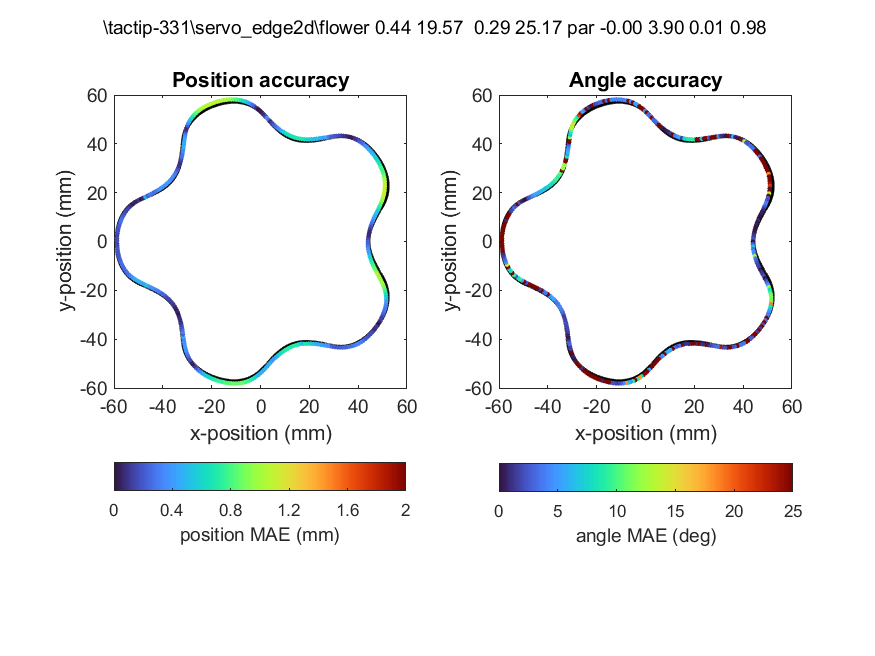} & 		
		\includegraphics[width=0.49\columnwidth,trim={28 50 36 35},clip]{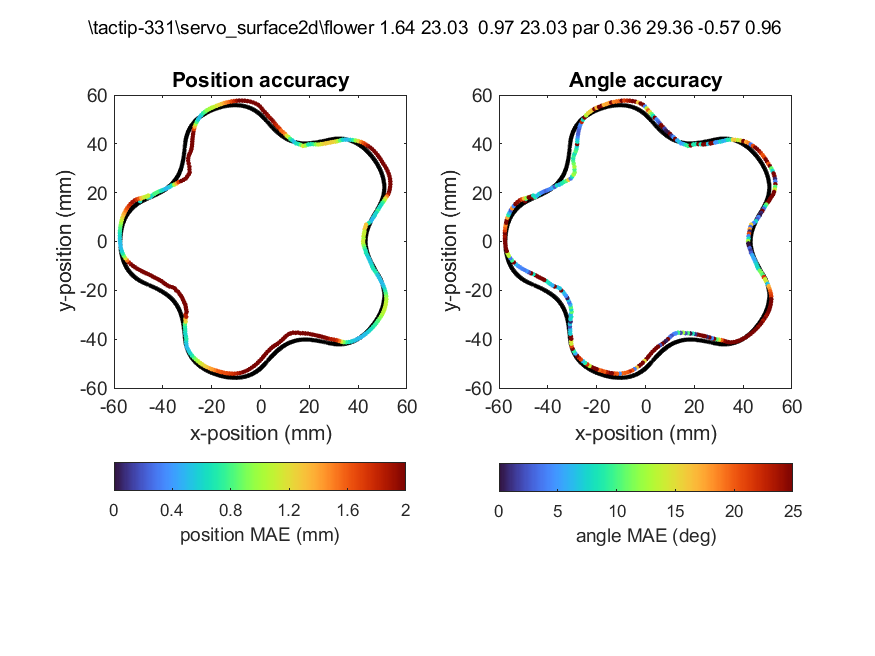} \\
	\end{tabular}
	\vspace{-1em}
	\caption{Edge and surface following with the DIGIT, DigiTac and TacTip for the disk, square and circular-wave shapes. }
	\vspace{-2em}
	\label{fig:9}
\end{figure}

When applied to edge following on the three test shapes, the DIGIT, DigiTac and TacTip all accurately traced around the shapes to sub-mm accuracy (Fig.~\ref{fig:9}, two left columns). For all three sensors and all three shapes, the mean absolute position errors were mainly sub-millimeter (Table~\ref{tab:4}, edge stimuli), with the TacTip the most accurate (0.4-0.5\,mm), followed by both the DIGIT and DigiTac (0.6-1.2\,mm). Overall, the circular wave is most demanding, followed by the square (because of the corners) and the circle is the easiest. All position errors are larger than the test predictions ($\sim\,$0.15\,mm), as expected because the servo control task introduces factors such as the corners on the square and curves of the circular-wave that differ from the straight edges used in training. 

The angle prediction performance showed more variability in how accurately the sensor maintains normality to the edge during servoing (Fig.~\ref{fig:9}; Table~\ref{tab:4}). For just the circle and square, the TacTip has the lowest angle error ($\lesssim$5$^\circ$), followed by both the DIGIT and DigiTac (10-15$^\circ$), \textcolor{black}{which we attribute to the larger domed TacTip behaving better under shear.} All sensors were inaccurate on the circular-wave (20-30$^\circ$), because of the difficulty of contour following around the shape.  

\begin{table}[t!]
	\caption{Accuracy of edge and surface following for the DIGIT, DigiTac and TacTip.}
	\vspace{0em}
	\begin{tabular}{@{}c|cccc@{}}
    	\textbf{sensor} & \textbf{stimulus} & \textbf{shape} & \textbf{position MAE} & \textbf{angle MAE} \\
    	\hline\hline
    	DIGIT & edge & circle & 0.6\,mm\ & 10.7$^\circ$ \\
    	 & edge & square & 0.9\,mm & 12.7$^\circ$ \\
    	 & edge & circular-wave & 1.2\,mm & 29.7$^\circ$ \\
    	 \hline
    	 DigiTac & edge & circle & 0.6\,mm\ & 15.1$^\circ$ \\
    	 & edge & square & 1.0\,mm & 15.0$^\circ$ \\
    	 & edge & circular-wave & 1.1\,mm & 26.2$^\circ$ \\
    	 & surface & circle & 0.7\,mm & 12.9$^\circ$ \\
    	 & surface & square & 1.5\,mm & 10.2$^\circ$ \\
    	 & surface & circular-wave & 1.0\,mm & 27.9$^\circ$ \\
    	 \hline
    	TacTip & edge & circle & 0.4\,mm & 1.6$^\circ$ \\
    	 & edge & square & 0.5\,mm & 4.6$^\circ$ \\
    	 & edge & circular-wave & 0.4\,mm & 19.6$^\circ$ \\
    	 & surface & circle & 0.6\,mm & 15.4$^\circ$ \\
    	 & surface & square & 1.5\,mm & 14.7$^\circ$ \\
    	 & surface & circular-wave & 1.6\,mm & 23.0$^\circ$ \\
    	\end{tabular}
	\label{tab:4}
	\vspace{0em}
\end{table}

\subsection{Sensor Comparison using Surface Following}
\label{sec:4b2}

The final comparison of the tactile sensors is on predicting surface angle and contact depth during sliding contacts, which is then used for tactile servo control around the vertical walls of the same three test objects used above.

For offline surface-pose prediction on the test data, both the DigiTac and TacTip accurately predict surface pose, with the DigiTac giving the best contact depth error (0.06\,mm) and angle error (0.5$^\circ$) (Table~\ref{tab:3}). These correspond to 1.5\% and $\sim$1\% of the ranges (4\,mm and 60$^\circ$), with little scatter on the prediction vs ground truth plots (Fig.~\ref{fig:8}). The accuracy in contact depth is similar to the repeatability (0.05\,mm) of the desktop robot, which may be a limiting factor. 

\textcolor{black}{For the DIGIT, the angle error (2.47$^\circ$) and contact depth error (0.22\,mm) are high compared to those for DigiTac and TacTip (Table~\ref{tab:3}). It was not possible to successfully follow a surface with this model. We attribute this difficulty as due to the much stiffer sensing surface, which allowed only a narrow (1\,mm) range of training depths and during surface following left little tolerance to errors during the servo control.}

When applied to surface following on the three test shapes, both the DigiTac and TacTip had similar accuracy (0.6-1.5\,mm), with the circle the most accurately traced at $\sim$0.5\,mm error (Fig.~\ref{fig:9}, Table~\ref{tab:4}). Both sensors struggled to turn the corners of the square, and for the DigiTac we needed to advance the angular set point to 5$^\circ$ for the task to complete. Like edge-following, the circular-wave was the most demanding shape to accurately trace. The angle errors were similar for the DigiTac/TacTip, being least for the circle and square \mbox{(10$^\circ$-15$^\circ$)} and largest on the circular-disk ($\sim$25$^\circ$). 

\section{Discussion}

\textcolor{black}{In this paper, we compared low-cost high-resolution tactile sensors from the GelSight and TacTip families, by adapting the DIGIT (GelSight) to have a TacTip sensing surface, The DIGIT-TacTip, or DigiTac allowed comparison of these two tactile sensor types with the same hardware and software.}


Overall, the TacTip, DigiTac and DIGIT optical tactile sensors all performed well at pose prediction and pose-based tactile servo control, with pose-prediction accuracy of $\sim\,$0.1\,mm and $\sim$1-2$^\circ$, and servo control accuracy typically better than 1\,mm. There was some biasing of contact angle during the servo control tasks, mainly for the DIGIT and DigiTac. \textcolor{black}{However, they still performed accurately at edge and surface following, except the DIGIT was not suited for sliding over surfaces because of its relatively flat and stiff elastomer.}

The most significant difference between the sensors is their material properties and construction. The DIGIT sensing surface is flat and fairly inelastic compared to the soft curved sensing surface of the TacTip. GelSight-type sensors have flat sensing surfaces comprised of a molded elastomer illuminated laterally, which are highly effective at imaging fine surface detail. In contrast, TacTip-type sensors come in various shapes and sizes~(Fig.~\ref{fig:2}) with a flexible 3D-printed skin and compliance from a soft gel. For the servo control tasks considered here, a soft curved tactile sensor was more practical to fit into curved surface and had greater tolerance for safe contact. 


The robustness of the sensors also depends on their material properties and construction, which is important because touch involves contact with the potential for damage. Our tests were demanding because: (i) thousands of sliding contacts were needed to train the pose-prediction models; (ii) surface and edge following shears the sensor surface as it rubs against the object; (iii) surface following can also lead to collisions with the test object. During the tests, we broke one DigiTac by ripping the skin where it joins the housing and one DIGIT by shearing off the elastomer from a collision. 

Because our aim in comparing tactile sensor technologies is to encourage innovation in the field, we will open-source the DigiTac sensing surface and software infrastructure including the robot interface and libraries for the experiments. We will also release all data and models used in this paper. As a final comment, we encourage others to adopt a similarly open perspective, and look forward to seeing further comparison and open release of other low-cost optical tactile sensors. 

{\em Acknowledgements:} We thank Mike Lambeta and Roberto Calandra for donating the DIGIT sensors, and Stephen Redmond for discussions including suggesting the name `DigiTac'.



\bibliographystyle{unsrt}
\bibliography{RAL-IROS2022.bib}

\begin{thebibliography}{10}

\bibitem{lambeta_digit_2020-2}
M.~Lambeta, P.~Chou, S.~Tian, B.~Yang, B.~Maloon, V.~Most, D.~Stroud,
  R.~Santos, A.~Byagowi, G.~Kammerer, D.~Jayaraman, and R.~Calandra.
\newblock {{DIGIT}}: {{A Novel Design}} for a {{Low-Cost Compact
  High-Resolution Tactile Sensor With Application}} to {{In-Hand
  Manipulation}}.
\newblock {\em IEEE Robotics and Automation Letters}, 5(3):3838--3845, July
  2020.

\bibitem{digit_notitle_nodate}
DIGIT.
\newblock http://www.digit.ml/.

\bibitem{lambeta_pytouch_2021}
M.~Lambeta, H.~Xu, J.~Xu, P.~Chou, S.~Wang, T.~Darrell, and R.~Calandra.
\newblock {{PyTouch}}: {{A Machine Learning Library}} for {{Touch Processing}}.
\newblock In {\em 2021 {{IEEE International Conference}} on {{Robotics}} and
  {{Automation}} ({{ICRA}})}, pages 13208--13214, {Xi'an, China}, May 2021.

\bibitem{wang_tacto_2022}
S.~Wang, M.~Lambeta, P.~Chou, and R.~Calandra.
\newblock {{TACTO}}: {{A Fast}}, {{Flexible}}, and {{Open-Source Simulator}}
  for {{High-Resolution Vision-Based Tactile Sensors}}.
\newblock {\em IEEE Robotics and Automation Letters}, 7(2):3930--3937, April
  2022.

\bibitem{noauthor_teaching_nodate}
Teaching robots to perceive, understand, and interact through touch.
\newblock
  https://ai.facebook.com/blog/teaching-robots-to-perceive-understand-and-interact-through-touch/.

\bibitem{ward-cherrier_tactip_2018-1}
B.~{Ward-Cherrier}, N.~Pestell, L.~Cramphorn, B.~Winstone, M.~E. Giannaccini,
  J.~Rossiter, and N.~Lepora.
\newblock The {{TacTip Family}}: {{Soft Optical Tactile Sensors}} with
  {{3D-Printed Biomimetic Morphologies}}.
\newblock {\em Soft Robotics}, 5(2):216--227, 2018.

\bibitem{lepora_soft_2021-1}
N.~Lepora.
\newblock Soft {{Biomimetic Optical Tactile Sensing With}} the {{TacTip}}: {{A
  Review}}.
\newblock {\em IEEE Sensors Journal}, 21(19):21131--21143, October 2021.

\bibitem{lepora_pixels_2019-1}
N.~Lepora, A.~Church, C.~de~Kerckhove, R.~Hadsell, and J.~Lloyd.
\newblock From {{Pixels}} to {{Percepts}}: {{Highly Robust Edge Perception}}
  and {{Contour Following Using Deep Learning}} and an {{Optical Biomimetic
  Tactile Sensor}}.
\newblock {\em IEEE Robotics and Automation Letters}, 4(2):2101--2107, 2019.

\bibitem{lepora_optimal_2020-1}
N.~Lepora and J.~Lloyd.
\newblock Optimal {{Deep Learning}} for {{Robot Touch}}: {{Training Accurate
  Pose Models}} of {{3D Surfaces}} and {{Edges}}.
\newblock {\em IEEE Robotics \& Automation Magazine}, 27(2):66--77, June 2020.

\bibitem{lepora_pose-based_2021}
N.~Lepora and J.~Lloyd.
\newblock Pose-{{Based Tactile Servoing}}: {{Controlled Soft Touch Using Deep
  Learning}}.
\newblock {\em IEEE Robotics \& Automation Magazine}, 28(4):43--55, December
  2021.

\bibitem{lepora_towards_2021}
N.~Lepora, A.~Stinchcombe, C.~Ford, A.~Brown, J.~Lloyd, M.~Catalano,
  M.~Bianchi, and B.~{Ward-Cherrier}.
\newblock Towards integrated tactile sensorimotor control in anthropomorphic
  soft robotic hands.
\newblock In {\em 2021 {{IEEE International Conference}} on {{Robotics}} and
  {{Automation}} ({{ICRA}})}, {Xi'an China}, May 2021.

\bibitem{johnson_retrographic_2009-1}
M.~K. Johnson and E.~H. Adelson.
\newblock Retrographic sensing for the measurement of surface texture and
  shape.
\newblock In {\em {{IEEE Conference}} on {{Computer Vision}} and {{Pattern
  Recognition}}}, pages 1070--1077, 2009.

\bibitem{yuan_gelsight_2017}
W.~Yuan, S.~Dong, and E.~Adelson.
\newblock {{GelSight}}: {{High-Resolution Robot Tactile Sensors}} for
  {{Estimating Geometry}} and {{Force}}.
\newblock {\em Sensors}, 17(12):2762, 2017.

\bibitem{gomes_geltip_2020-2}
D.~Gomes, Z.~Lin, and S.~Luo.
\newblock {{GelTip}}: {{A Finger-shaped Optical Tactile Sensor}} for {{Robotic
  Manipulation}}.
\newblock In {\em 2020 {{IEEE}}/{{RSJ International Conference}} on
  {{Intelligent Robots}} and {{Systems}} ({{IROS}})}, pages 9903--9909, {Las
  Vegas, NV, USA}, October 2020.

\bibitem{romero_soft_2020-1}
B.~Romero, F.~Veiga, and E.~Adelson.
\newblock Soft, {{Round}}, {{High Resolution Tactile Fingertip Sensors}} for
  {{Dexterous Robotic Manipulation}}.
\newblock In {\em 2020 {{IEEE International Conference}} on {{Robotics}} and
  {{Automation}} ({{ICRA}})}, pages 4796--4802, {Paris, France}, May 2020.

\bibitem{padmanabha_omnitact_2020-1}
A.~Padmanabha, F.~Ebert, S.~Tian, R.~Calandra, C.~Finn, and S.~Levine.
\newblock {{OmniTact}}: {{A Multi-Directional High-Resolution Touch Sensor}}.
\newblock In {\em 2020 {{IEEE International Conference}} on {{Robotics}} and
  {{Automation}} ({{ICRA}})}, pages 618--624, {Paris, France}, May 2020.

\bibitem{taylor_gelslim30_2022}
I.~Taylor, S.~Dong, and A.~Rodriguez.
\newblock {{GelSlim3}}.0: {{High-Resolution Measurement}} of {{Shape}},
  {{Force}} and {{Slip}} in a {{Compact Tactile-Sensing Finger}}.
\newblock In {\em 2022 {{IEEE International Conference}} on {{Robotics}} and
  {{Automation}} ({{ICRA}})}, pages 10781--10787, {Philadelphia, PA, USA}, May
  2022.

\bibitem{yuan_shape-independent_2017}
W.~Yuan, C.~Zhu, A.~Owens, M.~Srinivasan, and E.~Adelson.
\newblock Shape-independent hardness estimation using deep learning and a
  {{GelSight}} tactile sensor.
\newblock In {\em 2017 {{IEEE International Conference}} on {{Robotics}} and
  {{Automation}} ({{ICRA}})}, pages 951--958, {Singapore, Singapore}, May 2017.

\bibitem{luo_editorial_2021}
S.~Luo, N.~Lepora, U.~{Martinez-Hernandez}, Joao Bimbo, and H.~Liu.
\newblock Editorial: {{ViTac}}: {{Integrating Vision}} and {{Touch}} for
  {{Multimodal}} and {{Cross-Modal Perception}}.
\newblock {\em Frontiers in Robotics and AI}, 8, 2021.

\bibitem{chorley_development_2009-1}
C.~Chorley, C.~Melhuish, T.~Pipe, and J.~Rossiter.
\newblock Development of a tactile sensor based on biologically inspired edge
  encoding.
\newblock In {\em International {{Conference}} on {{Advanced Robotics}}}, pages
  1--6, 2009.

\bibitem{pestell_artificial_2022-1}
N.~Pestell, T.~Griffith, and N.~Lepora.
\newblock Artificial {{SA-I}} and {{RA-I}} afferents for tactile sensing of
  ridges and gratings.
\newblock {\em Journal of The Royal Society Interface}, 19(189):20210822, April
  2022.

\bibitem{lloyd_goal-driven_2022}
J.~Lloyd and N.~Lepora.
\newblock Goal-{{Driven Robotic Pushing Using Tactile}} and {{Proprioceptive
  Feedback}}.
\newblock {\em IEEE Transactions on Robotics}, 38(2):1201--1212, April 2022.

\bibitem{church_tactile_2021}
A.~Church, J.~Lloyd, R.~Hadsell, and N.~Lepora.
\newblock Tactile {{Sim-to-Real Policy Transfer}} via {{Real-to-Sim Image
  Translation}}.
\newblock In {\em Proceedings of the 5th {{Conference}} on {{Robot Learning}}},
  pages 1645--1654. {PMLR}, October 2021.

\bibitem{kamiyama_evaluation_2004}
K.~Kamiyama, H.~Kajimoto, Naoki Kawakami, and Susumu Tachi.
\newblock Evaluation of a vision-based tactile sensor.
\newblock In {\em {{IEEE International Conference}} on {{Robotics}} and
  {{Automation}} ({{ICRA}})}, pages 1542--1547 Vol.2, {New Orleans, LA, USA},
  2004.

\bibitem{zhang_deltact_2022}
G.~Zhang, Y.~Du, H.~Yu, and M.~Wang.
\newblock {{DelTact}}: {{A Vision-based Tactile Sensor Using Dense Color
  Pattern}}.
\newblock {\em arXiv preprint arXiv:2202.02179}, February 2022.

\bibitem{lin_sensing_2019}
X.~Lin and M.~Wiertlewski.
\newblock Sensing the {{Frictional State}} of a {{Robotic Skin}} via
  {{Subtractive Color Mixing}}.
\newblock {\em IEEE Robotics and Automation Letters}, 4(3):2386--2392, July
  2019.

\bibitem{li_elastomer-based_2019}
W.~Li, J.~Konstantinova, Y.~Noh, Z.~Ma, A.~Alomainy, and K.~Althoefer.
\newblock An {{Elastomer-based Flexible Optical Force}} and {{Tactile Sensor}}.
\newblock In {\em 2019 2nd {{IEEE International Conference}} on {{Soft
  Robotics}} ({{RoboSoft}})}, pages 361--366, {Seoul, Korea (South)}, April
  2019.

\bibitem{sferrazza_design_2019}
C.~Sferrazza and R.~D'Andrea.
\newblock Design, {{Motivation}} and {{Evaluation}} of a {{Full-Resolution
  Optical Tactile Sensor}}.
\newblock {\em Sensors}, 19(4):928, February 2019.

\end{thebibliography}

\end{document}